\documentclass{article}

\usepackage[main, final]{neurips_2026}

\usepackage[utf8]{inputenc} 
\usepackage[T1]{fontenc}    
\usepackage{hyperref}       
\usepackage{url}            
\usepackage{booktabs}       
\usepackage{amsfonts}       
\usepackage{nicefrac}       
\usepackage{microtype}      
\usepackage{xcolor}         
\usepackage{enumitem}
\usepackage{amsmath}
\usepackage{amssymb}
\usepackage{mathtools}
\usepackage{amsthm}
\usepackage[table]{xcolor}
\usepackage{array}
\usepackage{xspace}
\usepackage{multirow}
\usepackage{etoc}
\usepackage{graphicx}
\usepackage{float}
\usepackage{caption}
\usepackage{afterpage}
\etocsettocstyle{}{}
\usepackage[capitalize,noabbrev]{cleveref}
\usepackage{wrapfig}

\usepackage{subcaption}
\usepackage{tabularx}

\definecolor{cvprblue}{rgb}{0.21,0.49,0.74}
\definecolor{gaingreen}{rgb}{0.125,0.549,0.129}
\definecolor{darkgreen}{HTML}{083228}
\newcommand{\sname}{CAFT\xspace}

\newcommand{\snameOurs}{\sname (ours)\xspace}

\newcommand{\appx}[1]{Appendix~\ref{#1}}

\newcommand{\gain}[1]{\textbf{{\color{gaingreen}#1}}}
\newcommand{\treeforest}[1]{\textcolor{darkgreen}{#1}}

\newcommand{\ablationquestion}[2]{%
    \par\vspace{2pt}%
    \begingroup
    \setlength{\fboxsep}{4pt}%
    \noindent\colorbox{gray!10}{%
        \parbox{\dimexpr\linewidth-2\fboxsep\relax}{%
            \textbf{Q#1.}\quad\textbf{#2}%
        }%
    }%
    \par\endgroup
    \vspace{2pt}%
}

\newcolumntype{a}{>{\columncolor{gray!8}}c}

\newcommand{\sdag}{\textsuperscript{$\dagger$}}
\newcommand{\dagmark}{\rlap{\hspace{0.01em}\sdag}} 
\newcommand{\dagn}[1]{#1\dagmark}

\usepackage[textsize=tiny]{todonotes}
\usepackage{cleveref} 
\crefname{appendix}{Appendix}{Appendices}

\title{Aligning Forest and Trees in Images \& Long Captions \\ for Visually Grounded Understanding}

\author{
  Byeongju Woo$^{1,4}$ \quad
  Zilin Wang$^{1}$ \quad
  Byeonghyun Pak$^{2,4}$ \quad
  Sangwoo Mo$^{3}$ \quad
  Stella X. Yu$^{1}$ \\[0.55em]
  $^{1}$University of Michigan \quad
  $^{2}$University of Maryland \quad
  $^{3}$POSTECH \\
  $^{4}$Agency for Defense Development \\[0.20em]
  \texttt{\{byeongju,stellayu\}@umich.edu}
}

\begin{document}

\maketitle
\begingroup
\renewcommand\thefootnote{}
\footnotetext{This work was conducted using resources from the Agency for Defense Development (ADD).}
\addtocounter{footnote}{-1}
\endgroup


\def\figImTextHier#1{
\begin{figure*}[#1]
\centering
\includegraphics[width=0.9\textwidth]{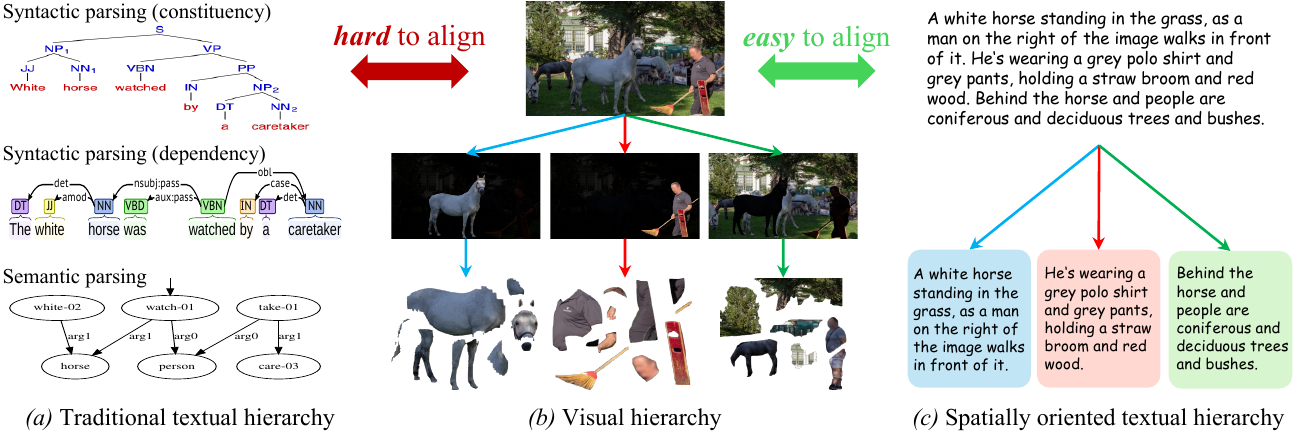}
\caption{
{\bf Vision-language understanding requires hierarchical semantics to be aligned across domains.} 
{\it Traditional textual hierarchies}, such as syntactic parse trees or semantic graphs, do not naturally correspond to the spatial organization of visual scenes.
In contrast, {\it spatially oriented textual hierarchies}, where each part describes a distinct image region, align naturally with {\it visual hierarchies}.
}
\label{fig:imTextHier}
\end{figure*}
}

\vspace{-3mm}

\begin{abstract}

Vision-language models such as CLIP often struggle to faithfully understand long, detail-rich captions, relying on dominant scene cues while overlooking fine-grained visual evidence.
We propose a hierarchical vision-language learning principle for understanding scenes as part-to-whole compositions: before forming a whole-scene representation, a model should uncover what semantic parts appear where in the image.
To this end, we propose \sname ({\it \underline{C}ross-domain \underline{A}lignment of \underline{F}orest and \underline{T}rees}), a vision-language model that jointly learns local text-region alignment at intermediate representations and global image-text alignment at the final representation.
Exploiting the organization of long captions, where local descriptions often correspond to scene parts, \sname employs a fine-to-coarse image encoder and a part-whole text encoder to discover localized part semantics and progressively compose them into a global image-text representation.
Trained on 30M image-text pairs, CAFT achieves state-of-the-art performance on six long-text retrieval benchmarks and exhibits strong scaling behavior. Experiments show that \sname learns fine-grained representations that localize textual semantics in image regions without explicit region-level supervision. 
\end{abstract}

\section{Introduction}

\begin{figure}[t]
    \centering
    \includegraphics[width=1.0\linewidth]{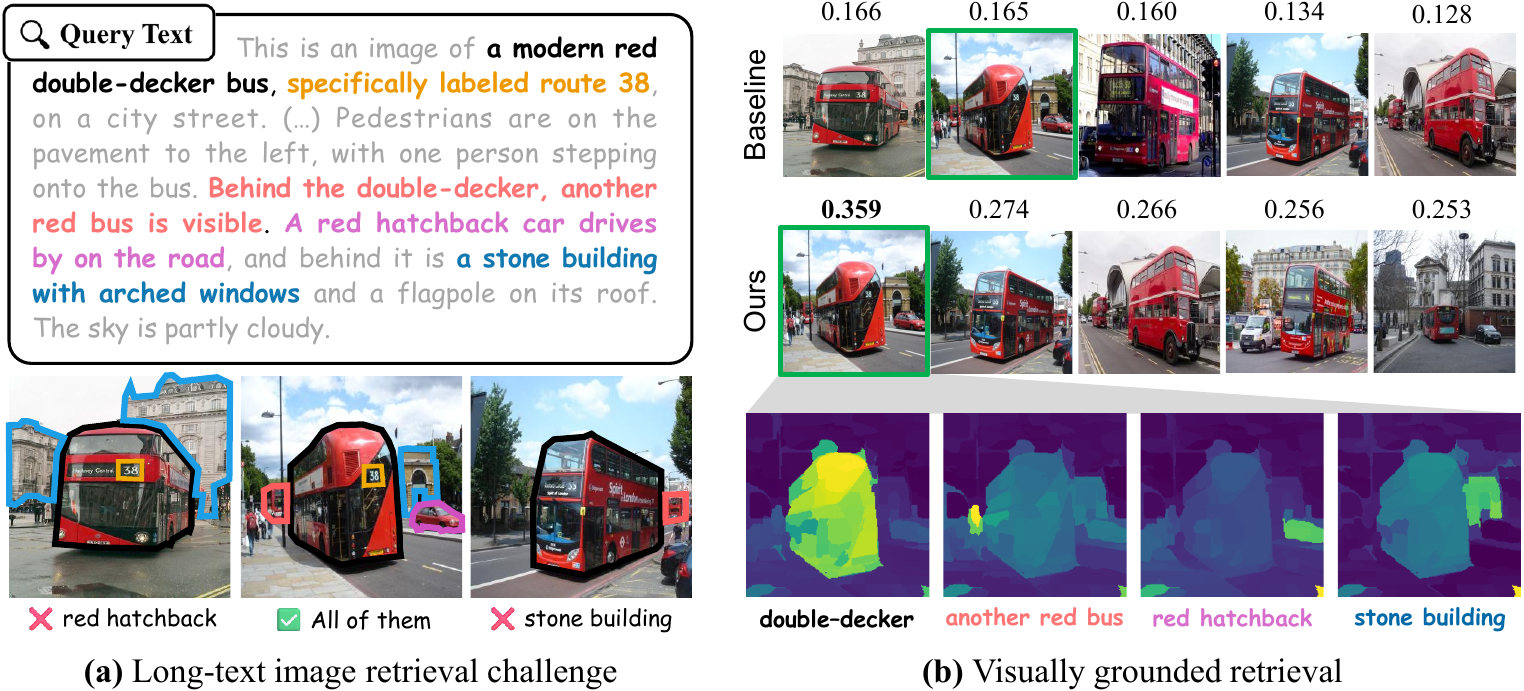}
    \caption{
        {\bf Fine-grained image-text retrieval requires composing local evidence into global understanding.}
        {\bf (a)} Among visually similar bus images, only the correct image contains all entities described in the long caption, including the {\it hatchback car} and the {\it stone building}. 
        Correct retrieval 
        requires identifying what appears where 
        and 
        integrating this local evidence into a whole-scene understanding.
        {\bf (b)} 
        Our model retrieves the correct image with a clearly higher matching score. 
        This result reflects its ability to ground individual textual descriptions in the corresponding image regions and compose these grounded part-level semantics into a global scene representation.
    }
    \label{fig:teaser}
    \vspace{-10pt}
    \phantomsubcaption\label{fig:teaser-a}
    \phantomsubcaption\label{fig:teaser-b}
    \vspace{-0.2mm}
\end{figure}

Vision-language models such as CLIP~\citep{pmlr-v139-radford21a} have demonstrated strong image-text understanding across diverse tasks. 
However, they often remain limited in capturing fine-grained visual entities and spatial relations described in detailed captions. 
This limitation is especially critical in long-caption image-text retrieval, where visually similar images must be distinguished using subtle local evidence.
As illustrated in \cref{fig:teaser-a}, a model may rely on a visually dominant cue, such as a red bus, while overlooking discriminative details, such as the labeled route number, or surrounding context, such as nearby cars and buildings~\citep{lavoie2026clip}. 
Successful retrieval therefore requires identifying what appears where in the image~\citep{lee2026pixellevel} and using this localized evidence to understand the full scene, rather than merely matching global image and text representations.

A line of work addresses fine-grained vision-language understanding by training models with long and detailed captions generated by MLLMs~\citep{zhang2024long,wu2024lotlip,najdenkoska2024tulip,chen2024sharegpt4v,zheng2024dreamlip}, which provide richer supervision by describing objects, attributes, and background context beyond concise image-text pairs. 
However, most methods still rely on global image-text alignment, which provides limited inductive bias for grounding each described entity and integrating it into the final scene representation~\citep{asokan2025finelip}.
Another line of work incorporates local visual-textual correspondences using privileged region-level supervision, such as bounding boxes~\citep{xie2025fg,choi2025goal}. 
While such supervision can improve local grounding, these approaches remain suboptimal for long-caption retrieval, as they do not explicitly encourage multiple localized details to be composed into a coherent whole-scene understanding.

In this paper, we propose a {\bf hierarchical vision-language learning principle} for fine-grained image-text understanding. 
Inspired by human perception, where visual scenes are naturally parsed into part-whole structures~\citep{hinton1979some}, we argue that whole-scene understanding should be built from grounded part-level semantics. 
Rather than learning only global image-text alignment, or applying part-level and whole-level alignment at the same representation level, a model should first learn part-level representations that capture local semantic entities and then compose them into a global scene representation. 
This part-to-whole inductive bias encourages the model to base whole-image understanding on localized visual evidence, reducing reliance on dominant cues and mitigating the tendency to overlook fine-grained details. 
As shown in \cref{fig:teaser-b}, this enables the model to retrieve the correct image by grounding individual textual descriptions in corresponding image regions and composing these grounded semantics into a whole-scene representation.

To realize this principle, we introduce {\bf \sname} ({\it \underline{C}ross-domain \underline{A}lignment of \underline{F}orests and \underline{T}rees}), a vision-language model for hierarchical image-text representation learning. 
\sname employs a {\it fine-to-coarse image encoder} and a {\it part-whole text encoder} to explicitly represent part-level components and progressively compose them into global representations within each modality. 
The fine-to-coarse image encoder discovers visual parts through progressive token grouping, while the part-whole text encoder encodes sub-captions as part-level textual units and aggregates them into a whole-caption representation. 
On top of these hierarchical representations, \sname aligns image and text at matched granularities: part-level visual features are aligned with sub-caption embeddings at intermediate layers, while whole-image and whole-caption representations are aligned at the final layer. 
This design promotes hierarchical representation composition, allowing local visual grounding to support global image-text understanding.

\begin{figure*}[!t]
    \centering
    \vspace{-4mm}
    \includegraphics[width=0.92\textwidth]{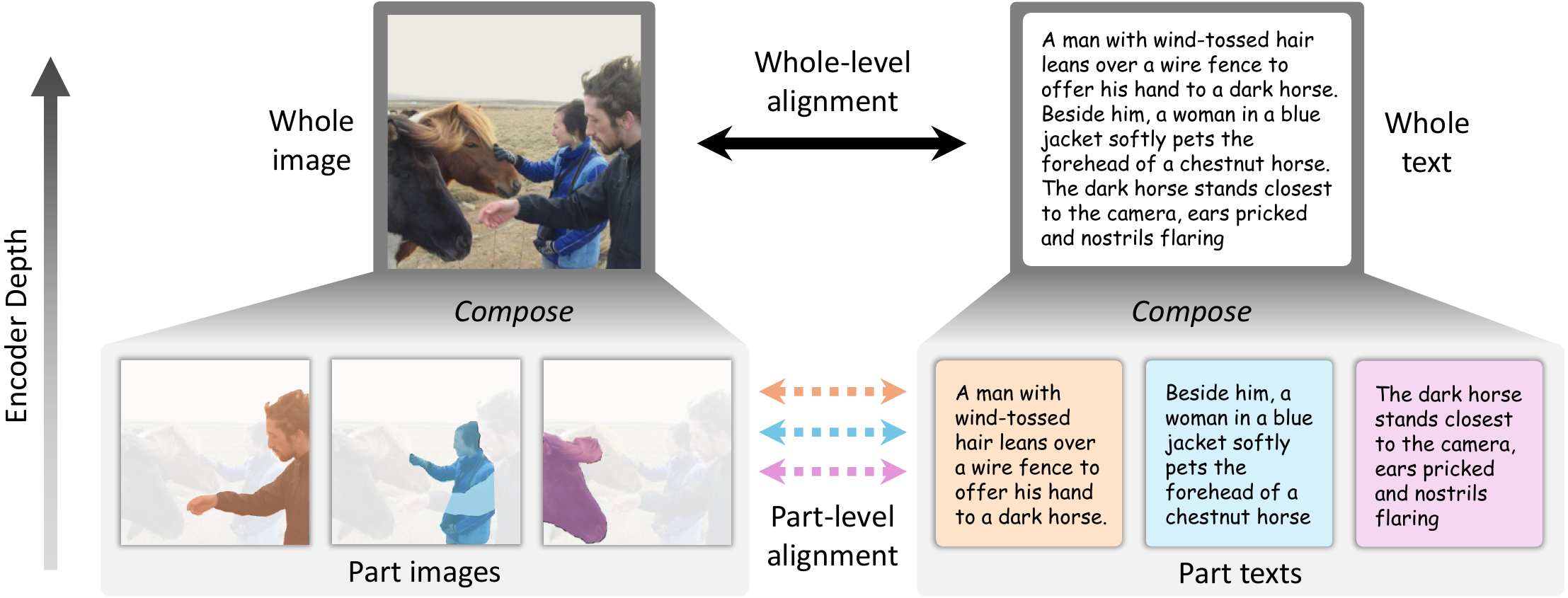}
    \caption{
        \textbf{Hierarchical vision-language learning.} 
        A correct understanding of the whole is established by understanding its constituent parts.
        We place part-level alignment
        (\textbf{\textit{\treeforest{trees}}})
        beneath whole-level alignment
        (\textbf{\textit{\treeforest{forest}}}),
        so that whole semantics are built from localized part semantics.
    }
    \label{fig:real_teaser}
    \vspace{-3mm}
\end{figure*}

Trained on 30M image-text pairs, CAFT achieves state-of-the-art performance across six long-caption retrieval benchmarks. 
CAFT consistently outperforms prior methods across data scales from 3M to 30M pairs, demonstrating strong scaling behavior and data efficiency.
Furthermore, we evaluate the learned part-level representations through zero-shot referring image segmentation. 
Without using any segmentation annotations, \sname outperforms representative methods such as GroupViT~\citep{xu2022groupvit} and MaskCLIP~\citep{zhou2022extract}, demonstrating that hierarchical image-text learning naturally gives rise to visually grounded part-level representations.

\newpage

Our contributions are summarized as follows:
\begin{itemize}[leftmargin=1.2em, itemsep=0.2em, topsep=0.2em]
    \item We propose a hierarchical vision-language learning principle for fine-grained image-text understanding, where whole-scene representations are composed from grounded part-level semantics.
    \item We introduce \sname, a vision-language model that implements this principle through a fine-to-coarse image encoder, a part-whole text encoder, and hierarchical cross-domain alignment.
    \item We demonstrate that \sname achieves state-of-the-art performance on six long-caption retrieval benchmarks and improves R@1 by up to 5.4\% over prior methods, while also learning strong visual grounding without privileged region-level annotations.
\end{itemize}

\vspace{-1mm}
Project page: \href{https://byeongju.me/CAFT/}{https://byeongju.me/CAFT/}.
\section{Related Work}

\textbf{CLIP with long-text capability.}
Long captions can describe detailed visual content in images, including fine-grained attributes and local details.
To exploit such captions, recent studies leverage synthetic long-caption datasets generated by multimodality large language models (MLLMs)~\citep{zheng2024dreamlip,chen2024sharegpt4v}.
Long-CLIP~\citep{zhang2024long} first extends pre-trained CLIP’s text context length from 77 to 248 tokens by stretching positional embeddings.
Subsequent studies improve this long-text adaptation~\citep{najdenkoska2024tulip,asokan2025finelip,xie2025fg}, while LoTLIP~\citep{wu2024lotlip} shows that long-text CLIP can also be trained from scratch.
However, longer captions and extended context length alone do not ensure that the model fully captures the rich information they provide.
For instance, \cite{lavoie2026clip} shows that the model often relies on the first sentence as a shortcut, overlooking fine-grained details in subsequent sentences.
In contrast, our part-whole text encoder explicitly encodes each part description and composes them into a full-caption embedding, better capturing the fine-grained information in long captions.

\textbf{Locality-aware vision-language pretraining.}
To improve fine-grained vision-language understanding~\citep{antol2015vqa,vinyals2015show,pak2024textual,wang2025oakCVPR}, prior work has explored local visual-textual alignment using various forms of region-text supervision.
GLIP~\citep{li2022grounded} learns phrase-region grounding from human-labeled noun phrase--bounding box pairs.
GOAL~\citep{choi2025goal} uses SAM~\citep{kirillov2023segment} to generate ROI bounding boxes and a CLIP to match them with sub-captions. 
FG-CLIP~\citep{xie2025fg} and PixCLIP~\citep{xiao2025pixclip} utilize MLLM-generated region descriptions paired with bounding boxes or segmentation masks.
Beyond explicit region-text labels, FILIP~\citep{yao2021filip} and FineLIP~\citep{asokan2025finelip} induce local alignment through token-wise maximum similarity, while FLAIR~\citep{xiao2025flair} learns to attend to text-relevant visual patches via text-conditioned attention pooling.

\vspace{-1mm}
However, these works mainly focus on local correspondences, while how the resulting regional understanding is properly reflected in the global understanding remains largely underexplored.
For example, FG-CLIP shows that adding regional image-text supervision improves region-level tasks, but does not lead to any meaningful gains in long-text understanding.
In contrast, our hierarchical learning principle aligns image and text at matched granularities with fine-to-coarse grouping, allowing aligned local semantics to be progressively composed into a locality-aware global representation.

\section{Cross-domain Alignment of Forest and Trees}
\label{sec:method}

We first present a general hierarchical vision-language learning principle for fine-grained image-text understanding~(\cref{sec:hierarchical-learning}). 
We then instantiate it as \sname with three components: a fine-to-coarse image encoder that forms a visual hierarchy~(\cref{sec:part-whole-structure-in-vision}), a part-whole text encoder that captures the textual hierarchy of long captions~(\cref{sec:part-whole-structure-in-language}), and a hierarchical alignment objective for whole- and part-level image-text alignment~(\cref{sec:part-to-whole-vision-languae-alignment}).

\subsection{Hierarchical vision-language learning principle}
\label{sec:hierarchical-learning}

Vision-language pretraining naturally aligns global image and text representations at the final encoder level to learn scene-level semantics.
A large body of work further incorporates region-text supervision, such as paired bounding boxes and regional descriptions, for local grounding, where these objectives are typically imposed at the same final level as global alignment.
Despite relying on additional regional supervision, this {\it flat alignment} can be suboptimal for fine-grained vision-language understanding because it treats local grounding and global alignment as parallel objectives rather than a compositional hierarchy.

%
We instead propose a {\bf hierarchical learning principle}: align local image-text semantics at intermediate representation levels, and global image-text semantics at the final level. 
This hierarchical placement of local and global objectives encourages global representations to be built from grounded local semantics, providing a bottom-up {\it hierarchical inductive bias}. 
The local objective encourages the model to identify \textit{what appears where} in the image (part), while the global objective composes these grounded cues into scene-level understanding (whole). 
As a result, global understanding is not learned solely from final-layer alignment, but is progressively supported by grounded intermediate representations, where spatially localized information is better preserved.

\begin{figure*}[!t]
    \centering
    \includegraphics[width=0.97\textwidth]{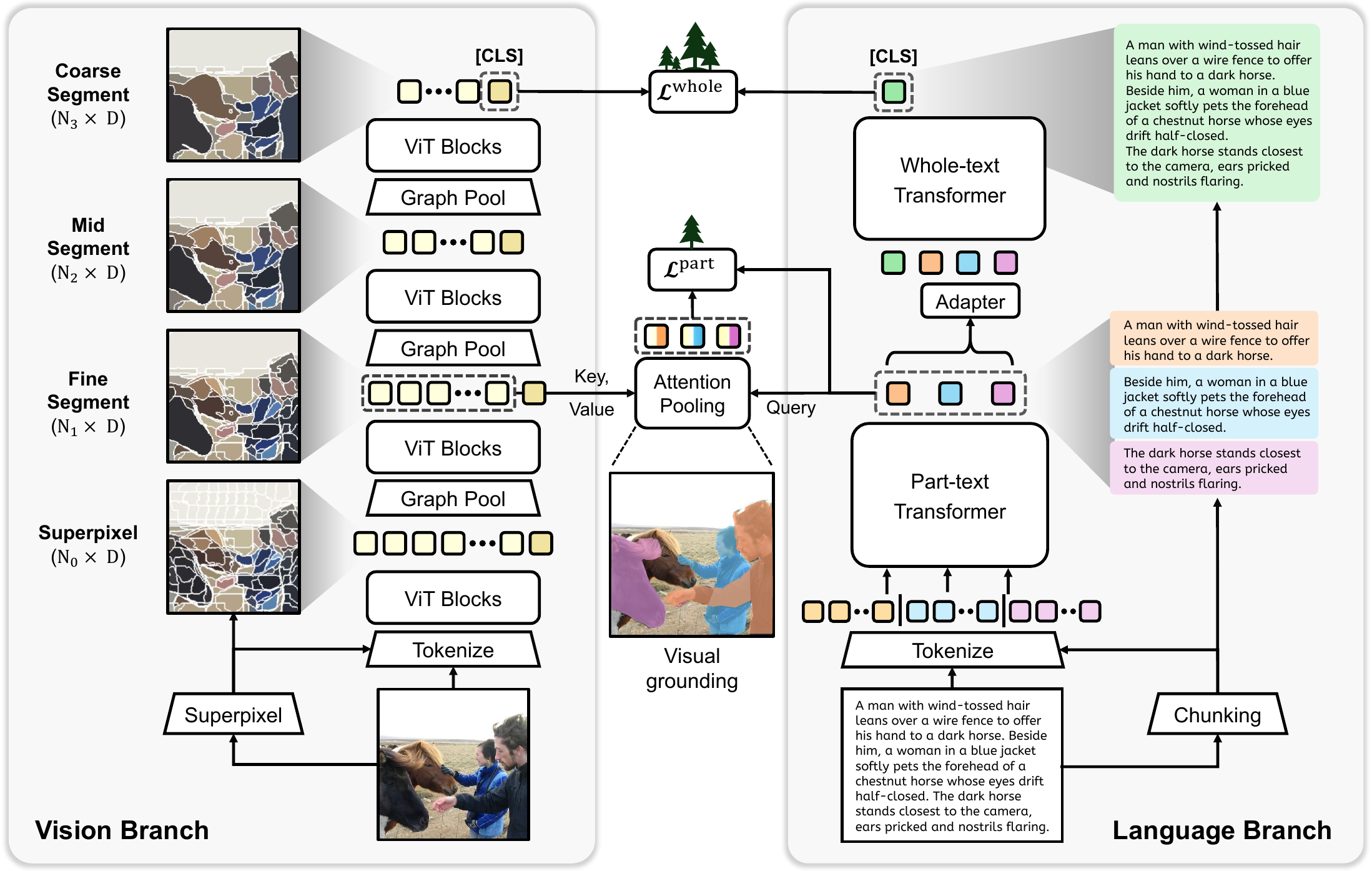}
    \caption{
        \textbf{Overview of \sname.}
        The model constructs hierarchical representations for both vision and language, aligning them at matched granularities.
        \textbf{Vision Branch}: Starting from superpixel tokens, the model performs fine-to-coarse scene parsing via progressive token grouping interleaved with ViT blocks.
        \textbf{Language Branch}: A Part-text Transformer encodes each part-text independently, followed by a Whole-text Transformer that aggregates them into a holistic embedding.
        \textbf{Alignment}: We establish a bottom-up hierarchy where intermediate visual features align with part-text embeddings for localized grounding (\textbf{\textit{\treeforest{aligning trees}}}), while final visual features align with whole-text embeddings to capture global scene semantics (\textbf{\textit{\treeforest{aligning forest}}}).
    }
    \label{fig:method}
    \vspace{-10pt}
\end{figure*}

\subsection{Fine-to-coarse image encoder}
\label{sec:part-whole-structure-in-vision}

{\bf Motivation.} 
To interpret a visual scene as a part-and-whole organization, a model should decompose the scene into meaningful visual units, such as objects and object sub-parts.
We posit that {\it explicitly} representing such units within the image encoder makes it easier to ground visual entities described in captions.
To this end, we build a {\bf fine-to-coarse image encoder}, which progressively groups visual tokens into meaningful segment tokens that represent object regions and their sub-parts.

{\bf Superpixel tokenization.} Instead of using uniform grid patches as in ViT, our encoder represents an image with superpixel tokens to better preserve object boundaries. See more details in \appx{supp:superpixel}.

{\bf Fine-to-coarse token grouping.} 
To implement progressive token grouping, we insert graph pooling between ViT blocks, following prior hierarchical segmentation architectures~\citep{ke2023cast}. 
Each graph pooling block merges tokens based on feature similarities, producing 64, 32, and 16 segment tokens from the initial 196 superpixel tokens. This yields a ViT-style encoder that explicitly discovers a visual hierarchy from fine-grained regions to coarse semantic parts.

Among these, the segment tokens in the second stage, denoted as $\mathbf{v}^{\mathrm{fine}} \in \mathbb{R}^{64 \times C}$, serve as intermediate patch representations for part-level alignment, while the \texttt{[CLS]} token from the final layer, denoted as $\mathbf{v}^{\mathrm{coarse}} \in \mathbb{R}^{C}$, serves as the final scene representation for whole-level alignment. 
This choice reflects that finer segment features better preserve object-level details, whereas coarser segment features better capture scene-level semantics~\citep{park2024visually}.

\subsection{Part-whole text encoder}
\label{sec:part-whole-structure-in-language}
{\bf Motivation.} 
We view the full-caption/sub-caption structure of long captions as a natural textual counterpart to the part-and-whole visual organization of a scene.
The full caption describes the entire scene, while its sub-captions often correspond to specific visual parts, including objects, object parts, and background context~\citep{zheng2024dreamlip}.
Under this view, we use sub-captions as part-texts to supervise local visual parts and the full caption as the whole-text to supervise the global scene representation.
To better capture this hierarchy, we propose a {\bf part-whole text encoder} that processes a long caption by independently encoding its part-texts into distinct part-text embeddings and then aggregating them into a whole-text embedding.


\textbf{Part-text chunking.}
We first obtain part-text by splitting the long caption into $N$ chunks $T_1, T_2, ..., T_N$: we split the original caption into individual sentences, and then concatenate 1-3 consecutive sentences into each chunk. See more details in the \appx{sec:chunking}.

\textbf{Part-text Transformer.}
Next, we {independently} forward each chunk through Part-text Transformer to encode $N$ part-text embeddings $\mathbf{t}^{\text{sub}}$. 
\begin{equation}
\label{eq:part_description_embed}
\mathbf{t}^{\text{part}}_{k} = \mathrm{Transformer}_{\text{part}}\!\left(\mathrm{Tokenize}(T_k)\right)\in \mathbb{R}^{D},
\qquad
\mathbf{t}^{\text{part}} = \{ \mathbf{t}^{\text{part}}_{k} \}_{k=1}^{N} \in \mathbb{R}^{N \times D}.
\end{equation}
\textbf{Whole-text Transformer.}
Finally, we obtain the whole-text embedding $\mathbf{t}^{\text{whole}} \in \mathbb{R}^{D}$ by feeding the adapted part-text embeddings together with a  \texttt{[CLS]} token into the Whole-text Transformer. 
The adaptation is implemented with a light-weight residual MLP adapter~\citep{gao2024clip}
\begin{equation}
\label{eq:whole_description_embed}
\mathbf{t}^{\text{whole}} = \mathrm{Transformer}_{\text{whole}}\!\left(\mathbf{\tilde{t}}^{\text{part}};\texttt{[CLS]}]\right) \in \mathbb{R}^{D},
\qquad
\mathbf{\tilde{t}}^{\text{part}}
= \mathbf{t}^{\text{part}} + \mathrm{MLP}(\mathbf{t}^{\text{part}}),
\end{equation}

\subsection{Hierarchical vision-language alignment loss}
\label{sec:part-to-whole-vision-languae-alignment}

\textbf{Training batch construction}. We train \sname from scratch on a synthetic long text-image paired dataset, where each image is annotated with multiple long and short captions. At each iteration, we construct a batch of $B$ images, each image $I_i$ paired with $K$ positive sub-captions $\{ T_{i_k} \}_{k=1}^K$.
Among the $K$ sub-captions, the first $N$ sub-captions are derived from a single long caption and are used as the input to the Whole-text Transformer, while the remaining ($K$$-$$N$) sub-captions are sampled from other captions. 
The resulting batch contains input pairs $\{ (I_i , \{ T_{i_k} \}_{k=1}^K ) \}_{i=1}^B$, intermediate features $\{ (\mathbf{v}^{\text{fine}}_{i} , \{ \mathbf{t}^{\text{part}}_{i_k} \}_{k=1}^K ) \}_{i=1}^B$, and final features $\{ (\mathbf{v}^{\text{coarse}}_{i} , \mathbf{t}^{\text{whole}}_{i} ) \}_{i=1}^B$.


\textbf{Training objective.}
Our objective consists of two alignment losses, $\mathcal{L} = \mathcal{L}^{\text{part}} + \mathcal{L}^{\text{whole}}$.
The whole-level loss $\mathcal{L}^{\text{whole}}$ aligns the entire image with the long caption at the final layer, while the part-level loss $\mathcal{L}^{\text{part}}$ aligns each part-text with its corresponding visual region at the intermediate layer.
We use sigmoid losses instead of softmax losses, which are more stable with small batch sizes and naturally supports multiple positives~\citep{zhai2023sigmoid}.
For both losses, $y_{i,j}$ indicates whether image $i$ and text $j$ form a positive or negative pair, with $y_{i,j}=+1$ for positive pairs and $y_{i,j}=-1$ for negative pairs.

\textbf{Whole-level Sigmoid Loss} 
aligns the global image feature $\mathbf{v}^{\text{coarse}}_{i}$ with the whole text feature $\mathbf{t}^{\text{whole}}_{i}$, as in standard CLIP training. 
Here, $\langle \cdot, \cdot \rangle$ denotes cosine similarity and $\sigma(\cdot)$ is the sigmoid function.
We use a learnable temperature $\tau_{\text{w}}$ and bias $b_{\text{w}}$ for the whole-level loss:
\begin{equation}
\label{whole_loss}
\mathcal{L}^{\text{whole}}_{i,j}
= -\log\,\sigma\!\left(
y_{i,j}\bigl(\tau_{\text{w}}\langle \mathbf{v}^{\text{coarse}}_{i},\,\mathbf{t}^{\text{whole}}_{j}\rangle + b_{\text{w}}\bigr)
\right).
\end{equation}

\textbf{Part-level Text-grounded Sigmoid Loss}
aligns each part-text feature with its corresponding visual region features without dense annotation (e.g., bounding box).
To this end, we use an attention pooling module~\citep{lavoie2024modeling,xiao2025flair} to obtain text-grounded visual features.
Let $\mathbf{v}^{\text{fine}}_{i}$ denote visual segment features from image $i$, and $\mathbf{t}^{\text{part}}_{j_k}$ denote $k$-th part-text features from caption $j$. Then attention pooling computes a \textbf{t}ext-\textbf{g}rounded visual feature $\mathbf{v}^{\text{tg}}_{i,j_k} = \mathrm{AttnPool}\!\left(\mathbf{t}^{\text{part}}_{j_k},\, \mathbf{v}^{\text{fine}}_{i}, \mathbf{v}^{\text{fine}}_{i}\right) \in \mathbb{R}^{D}$ by aggregating visual segments relevant to the part-text query.
We then compare $\mathbf{v}^{\text{tg}}_{i,j_k}$ with the query $\mathbf{t}^{\text{part}}_{j_k}$ using separate learnable parameters $\tau_{\text{p}}$ and $b_{\text{p}}$:
\begin{equation}
\label{part_loss}
\mathcal{L}^{\text{part}}_{i,j,k}
= -\log\,\sigma\!\left(
y_{i,j}\bigl(\tau_{\text{p}}\langle \mathbf{v}^{\text{tg}}_{i,j_k},\,\mathbf{t}^{\text{part}}_{j_k}\rangle+b_{\text{p}}\bigr)
\right).
\end{equation} 

\textbf{Inference.} 
At inference time, CAFT computes the image-text alignment score as a weighted sum of whole-level and part-level similarities. We set $\alpha=0.3$ by default and analyze its effect in \cref{tab:ablationPartScore}:
\begin{equation}
\label{eq:weighted_sum}
s_{ij}
= (1-\alpha)\cdot\langle \mathbf{v}^{\text{coarse}}_{i},\,\mathbf{t}^{\text{whole}}_{j}\rangle
+ \alpha\cdot\Sigma_k\langle \mathbf{v}^{\text{tg}}_{i,j_k},\,\mathbf{t}^{\text{part}}_{jk}\rangle
\end{equation} 

\section{Experiments}
\label{sec:exp}
To assess the fine-grained vision-language understanding of \textbf{CAFT}, we first evaluate image-text retrieval on long-caption benchmarks~(\cref{sec:benchmark}).
We then examine the visual grounding of the learned part-level representations that underlie holistic image-text understanding~(\cref{sec:vis-grounding}).
Finally, we conduct ablation studies to evaluate the contributions of individual components and compare different alignment strategies within our hierarchical learning framework~(\cref{sec:ablation}).


\subsection{Setup}
\label{sec:exp-setup}
\textbf{Training Datasets.}  We train \sname on the re-captioned datasets from DreamLIP~\citep{zheng2024dreamlip}, which provide long, detailed captions. 
This dataset consists of CC3M-recap, CC12M-recap, and YFCC15M-recap, which are aggregated into a unified 30M-sample corpus (Merged-30M).

\textbf{Implementation Details.}
Our vision encoder has a size comparable to ViT-B/16~\citep{dosovitskiy2020image} and uses a 224 $\times$ 224 input resolution.
For the text encoder, we follow the vanilla Transformer architecture~\citep{vaswani2017attention} with a two-stage design: a Part-text Transformer ($L_1=8$) and a Whole-text Transformer ($L_2=4$).
The total depth ($L=12$) matches the original CLIP text encoder~\citep{pmlr-v139-radford21a}.
We use 512-dimensional image and text embeddings, a sub-caption context length of 77, a whole-caption context length of $N=4$, and $K=8$ sub-captions per image.
We adopt the training hyperparameters of \cite{xiao2025flair}, but reduce the batch size to 2K due to GPU memory constraints.
Further details are provided in \appx{supp:exp_setup}.


\begin{table*}[!t]
\centering
\setlength{\tabcolsep}{5pt}
\caption{%
    \textbf{\sname achieves state-of-the-art results on zero-shot long text-image retrieval.}
    I2T and T2I denote image-to-text and text-to-image R@1, respectively. 
    The best results are \textbf{bold}, second-best are \underline{underlined}, and $\dagger$ denotes use of bounding box annotations. 
}
\vspace{1mm}
\resizebox{\textwidth}{!}{
\begin{tabular}{l l | aa cc | aa cc aa cc}
\toprule
\multirow{2}{*}{Method} 
& \multirow{2}{*}{Data}
& \multicolumn{2}{a}{DCI}
& \multicolumn{2}{c|}{DOCCI}
& \multicolumn{2}{a}{SV-1k}
& \multicolumn{2}{c}{SV-10k}
& \multicolumn{2}{a}{Urban-1k} 
& \multicolumn{2}{c}{IIW} \\
&
& I2T & T2I & I2T & T2I & I2T & T2I & I2T & T2I & I2T & T2I & I2T & T2I \\
\midrule

\multicolumn{6}{l}{\textit{Trained on Short-Captions Only}} \\[3pt]
OpenCLIP & 2B  & 56.0 & 55.4 & - & - & 90.3 & 87.7 & 69.6 & 66.8 & 69.5 & 65.8&  -  & - \\
SigLIP     & 10B & 57.7 & 56.2 & - & - & 85.8 & 83.4 & 83.4 & 63.0 & 62.7 & 62.1 &  -  & - \\
\midrule
\multicolumn{6}{l}{\textit{Trained on Short Captions $\to$ Finetuned on Long-Captions}} \\[3pt]
Long-CLIP  & 400M$\to$1M & 51.6 & 57.1 & 63.1 & 71.4 & 90.6 & 87.4 & 73.1 & 62.0 & 78.9 & 79.5 &  -  & -\\
TULIP  & 400M$\to$1M & - & - & - & - & \underline{98.6} & \underline{98.6} & - & - & 88.1 & 86.6 &  -  & - \\
FineLIP  & 400M$\to$1M & - & - & \underline{80.0} & {78.1} & - & - & - & - & {{91.2}} & {90.0} &  -  & - \\
DeBias-CLIP  & 400M$\to$1M & \underline{68.5} & \underline{67.6} & {79.7} & \underline{80.0} & - & - & - & - & \underline{93.1} & \underline{93.0} &  -  & - \\
\dagn{FG-CLIP}  & 400M$\to$1.6B & 61.8 & 60.6 & - & - & 96.7 & 94.9 & - & - & - & - &  -  & - \\
\dagn{FG-CLIP 2}  & 10B$\to$1.6B & 64.5 & 64.9 & - & - & 95.8 & 95.4 & - & - & - & - &  -  & - \\
\midrule
\multicolumn{6}{l}{\textit{Trained on Long-Captions from Scratch}} \\[3pt]
LoTLIP     & 100M & {62.1} & 61.0 & - & - & 95.5 & 86.8 & 86.8 & 81.4 &  88.8  &  84.8 & \underline{94.0} & \underline{92.5}  \\
FLAIR           & 30M & 61.3 & {66.2} & 70.3 & 72.6 & 
               98.5 & 98.0 & \underline{90.3} & \underline{89.4} & 83.6 & 87.7&  91.3 & 91.5 \\
\textbf{\snameOurs}  
& 30M & \textbf{69.7} &\textbf{72.4} & \textbf{80.4} & \textbf{82.0} & \textbf{99.5} & \textbf{99.0} & \textbf{95.5} &  \textbf{94.8} & \textbf{93.6} & \textbf{95.4} & \textbf{97.6} &\textbf{97.4} \\
\gain{\textit{vs. previous SOTA}} &  & \gain{+1.2} & \gain{+4.8} & \gain{+0.4} & \gain{+2.0} & \gain{+0.9} & \gain{+0.4} & \gain{+5.2} &  \gain{+5.4} & \gain{+0.5} & \gain{+2.4} & \gain{+3.6} & \gain{+4.9} \\
\bottomrule
\end{tabular}
}
\label{tab:long_text_image_retrieval}
\vspace{-3pt}
\end{table*}
\begin{figure*}[!t]
\centering

\includegraphics[width=1\linewidth]{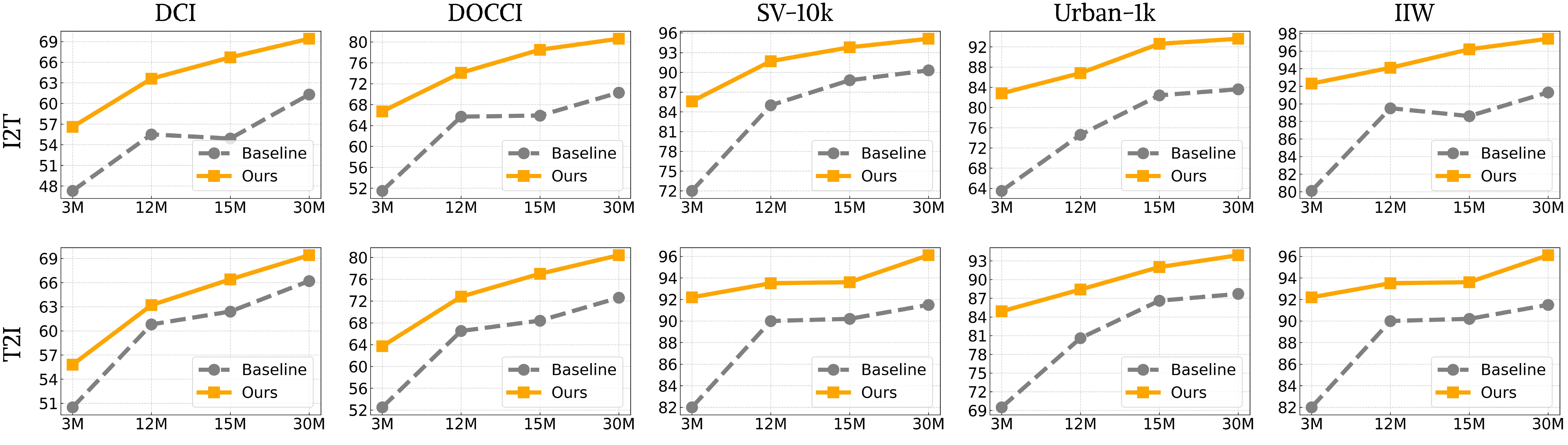} 

\caption{%
\textbf{\sname exhibits strong scaling behavior.}
We evaluate how \sname scales with training data by comparing it with FLAIR, a strong competitor trained on the same datasets, across training set sizes of 3M, 12M, 15M, and 30M.
\sname achieves consistent improvements as the training set grows, highlighting its robust scaling behavior and the effectiveness of its hierarchical inductive bias.
}

\vspace{-15pt}
\label{fig:flair_vs_ours}
\end{figure*}


\subsection{Image-Text Retrieval}
\label{sec:benchmark}

\textbf{Evaluation Benchmark.} We evaluate Recall@1 (R@1) for long text-image cross-modal retrieval on six widely used benchmarks: DCI~\citep{urbanek2024picture}, DOCCI~\citep{onoe2024docci}, ShareGPT4V-1k~\citep{chen2024sharegpt4v}, ShareGPT4V-10k~\citep{chen2024sharegpt4v}, Urban-1k~\citep{zhang2024long}, and IIW~\citep{garg2024imageinwords}. Detailed statistics for each dataset are provided in \appx{sec:suppl_benchmark}. 

\textbf{Comparison with State-of-the-art.}
We group prior work into three settings: (i) training only on short captions, (ii) pre-training on short captions followed by long-caption fine-tuning, and (iii) training from scratch on long captions.
\cref{tab:long_text_image_retrieval} compares models across six long-caption benchmarks, exceeding the two or three datasets typically used in prior studies.
Across these settings, \sname achieves state-of-the-art performance, substantially outperforming FLAIR at the same data scale (30M pairs) and surpassing fine-tuned models with far fewer pre-training samples.
These results demonstrate the effectiveness and data efficiency of our hierarchical design for long-caption retrieval.

\textbf{Scaling Behavior.}
Across scales from 3M to 30M pairs, \sname consistently outperforms FLAIR with a similar scaling slope (\cref{fig:flair_vs_ours}).
These results demonstrate that our hierarchical inductive bias supports robust scaling behavior and provides strong data efficiency.

\textbf{Short-text Image Retrieval.}
We further evaluate CAFT on standard short-text retrieval benchmarks in \appx{supp:short_text}.
CAFT achieves 67.1 I2T and 52.6 T2I on the MSCOCO~\citep{lin2014microsoft} validation split, demonstrating its effectiveness beyond long-text retrieval scenarios.



\vspace{-2pt}
\subsection{Visual Grounding}
\label{sec:vis-grounding}
\vspace{-2pt}

Visual grounding is the ability to localize image regions corresponding to a textual description~\citep{plummer2015flickr30k}.
In this section, we investigate how well \sname grounds sub-captions to their corresponding regions, supporting fine-grained recognition.

\begin{figure}[!t]
\centering

\includegraphics[width=1\columnwidth]{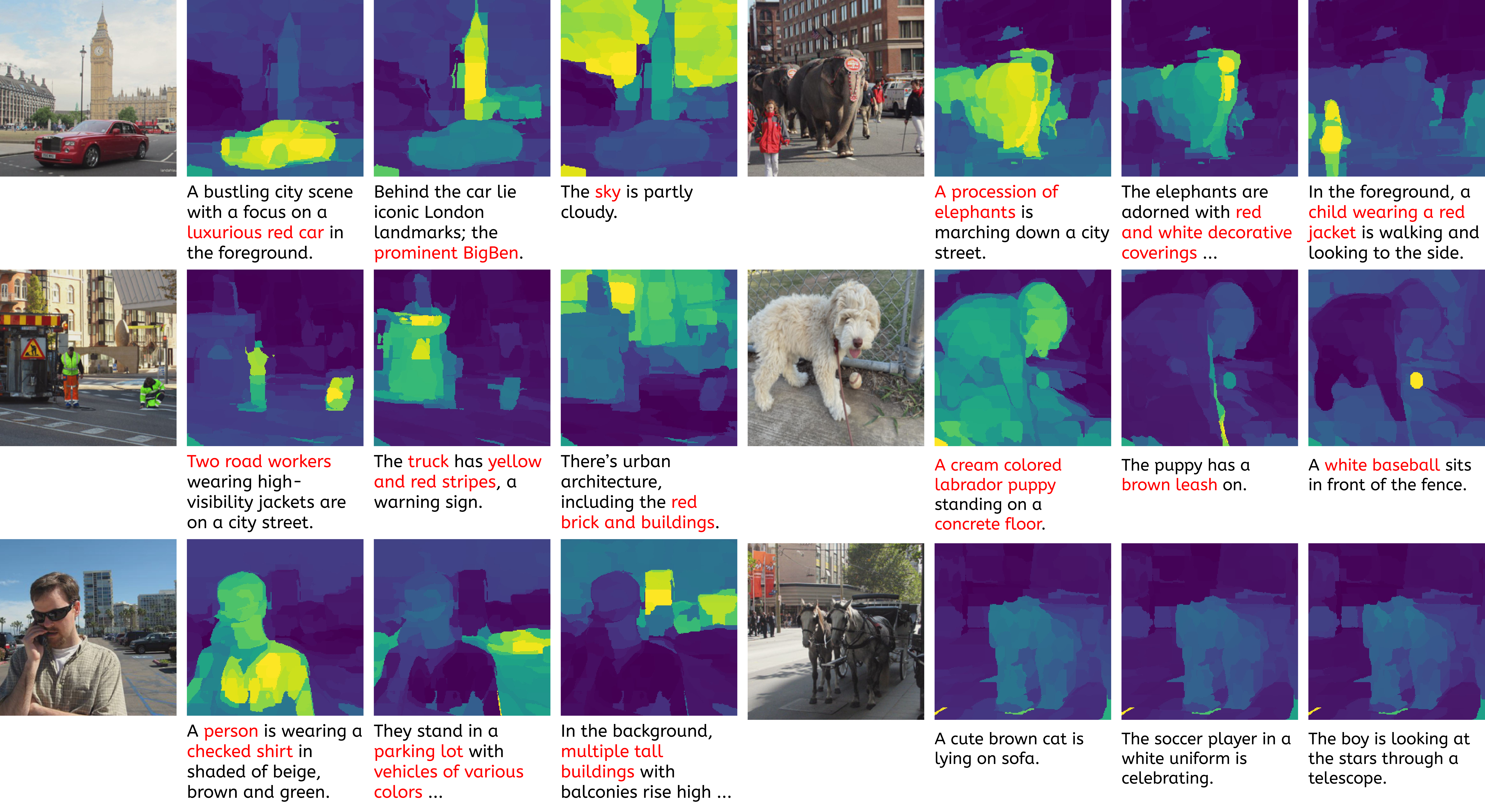}

\caption{
    \textbf{\sname provides spatially precise grounding.}
    We show head-averaged attention maps from the $\mathrm{AttnPool}$ module for each sub-caption.
    \sname precisely localizes relevant regions across different scales and positions, with sharp, object-aware boundaries. 
    The maps also exhibit inactivation for irrelevant captions, demonstrating robustness to mismatched semantic cues (last example).
}
\vspace{-15pt}
\label{fig:text_attn_map}
\end{figure}

\textbf{Qualitative Results.}
We visualize head-averaged attention maps from the $\mathrm{AttnPool}$ module in \cref{fig:text_attn_map} to examine how \sname localizes sub-captions.
As shown in the first example, \sname accurately grounds both salient objects and background context, such as the {\it car} and {\it sky}.
It also captures compositional relations in the phrase ``\textit{Behind the car lies the prominent Big Ben},'' attending to both entities while correctly prioritizing the subject (Big Ben) over the spatial reference (car).
The example in the top row of the second column further shows that \sname captures multiple levels of granularity, grounding both whole objects (e.g., elephants) and fine-grained details (e.g., decorative coverings).
Further examples are available in \appx{sec:suppl_more_vis}.

\begin{wraptable}[11]{r}{0.57\textwidth}
    \centering
    \small
    \setlength{\tabcolsep}{4.8pt}
    \renewcommand{\arraystretch}{0.95}
    
    \vspace{-1pt}
    \caption{%
        \textbf{Zero-shot referring image segmentation.}
        We report mIoU for models trained without mask supervision.
        CAFT achieves the best performance across all splits.
    }
    \label{tab:ris}
    \begin{tabular}{@{}l | ccc | ccc | c@{}}
        \toprule
        \multirow{2}{*}{Method}
        & \multicolumn{3}{c|}{RefCOCO} 
        & \multicolumn{3}{c|}{RefCOCO+} 
        & GRef \\
        & val & testA & testB 
        & val & testA & testB 
        & val \\
        \midrule
        GViT  & 8.0  & 6.2  & 10.5 & 8.5  & 6.8  & 10.6 & 10.7 \\
        MCLIP & 11.5 & 11.9 & 12.1 & 11.9 & 12.0 & 12.6 & 12.7 \\
        SaG   & 21.8 & 19.0 & 25.0 & 22.2 & 19.9 & 24.9 & 25.9 \\
        FLAIR & 19.8 & 19.8 & 20.0 & 18.5 & 18.3 & 18.9 & 20.2 \\
        \rowcolor{gray!10}
        \sname & \textbf{28.9} & \textbf{30.8} & \textbf{28.5}
               & \textbf{28.5} & \textbf{29.8} & \textbf{28.3}
               & \textbf{31.4} \\
        \bottomrule
    \end{tabular}
\end{wraptable}

\vspace{-10pt}
\textbf{Quantitative Results.}
We further evaluate this grounding behavior through zero-shot referring image segmentation.
Given a target image and a referring text description, we leverage the head-averaged attention maps and apply Otsu thresholding~\citep{otsu1979threshold} to obtain binary segmentation masks.
Following the evaluation protocol of SaG~\citep{kim2023shatter}, we report mIoU on the RefCOCO, RefCOCO+~\citep{yu2016modeling}, and GRef~\citep{mao2016generation}.

We compare \sname with prior approaches trained solely on image-text pairs.
As shown in \cref{tab:ris}, \sname consistently outperforms representative zero-shot methods such as GroupViT~\citep{xu2022groupvit} and MaskCLIP~\citep{zhou2022extract}.
Compared to FLAIR, which applies $\mathrm{AttnPool}$ over ViT patch tokens, \sname shows substantially stronger grounding.
This suggests that meaningful segment tokens uncovered by the fine-to-coarse encoder provide stronger grounding cues than standard ViT patch tokens.
Overall, the results indicate that \sname learns grounded representations that align textual descriptions with spatial regions, without any segmentation supervision.


\section{Ablation Studies}
\label{sec:ablation}

We organize our ablation studies around three key questions to examine our hierarchical vision-language learning principle and the design choices that support it.
All experiments are conducted on CC3M-recap.


\subsection{Hierarchical vs. Flat Alignment}
\label{sec:ablation_strategies}

\ablationquestion{1}{What does our hierarchical learning principle offer beyond flat alignment?}

\begin{figure}[!t]
    \centering
    {\small
    \setlength{\tabcolsep}{5.0pt}
    \begin{tabular}{@{}c l c cc cc cc cc cc@{}}
        \toprule
        \multirow{2}{*}{ }
        & \multirow{2}{*}{Method}
        & \multirow{2}{*}{Average}
        & \multicolumn{2}{c}{DCI}
        & \multicolumn{2}{c}{DOCCI}
        & \multicolumn{2}{c}{SV-10k}
        & \multicolumn{2}{c}{Urban-1k}
        & \multicolumn{2}{c}{IIW} \\
        \cmidrule(lr){4-5}
        \cmidrule(lr){6-7}
        \cmidrule(lr){8-9}
        \cmidrule(lr){10-11}
        \cmidrule(lr){12-13}
        & &
        & I2T & T2I
        & I2T & T2I
        & I2T & T2I
        & I2T & T2I
        & I2T & T2I \\
        \midrule
        {\bf (a)} & Whole
        & 72.5
        & 53.3 & 49.9
        & 63.9 & 57.8
        & 80.9 & 80.8
        & 78.4 & 81.2
        & 90.7 & 87.9 \\
        {\bf (b)} & Part + Whole
        & 72.6
        & 53.7 & 52.7
        & 62.5 & 60.1
        & 82.0 & 81.2
        & 74.5 & 80.0
        & 90.5 & 89.1 \\
        \rowcolor{gray!10}
        {\bf (c)} & Part $\rightarrow$ Whole\phantom{0}\phantom{0}
        & \textbf{78.1}
        & \textbf{57.6} & \textbf{59.6}
        & \textbf{67.3} & \textbf{67.1}
        & \textbf{86.2} & \textbf{85.4}
        & \textbf{83.6} & \textbf{87.2}
        & \textbf{93.6} & \textbf{93.3} \\
        \bottomrule
    \end{tabular}
    }
    \par\vspace{2.2mm}
    \includegraphics[width=0.90\linewidth]{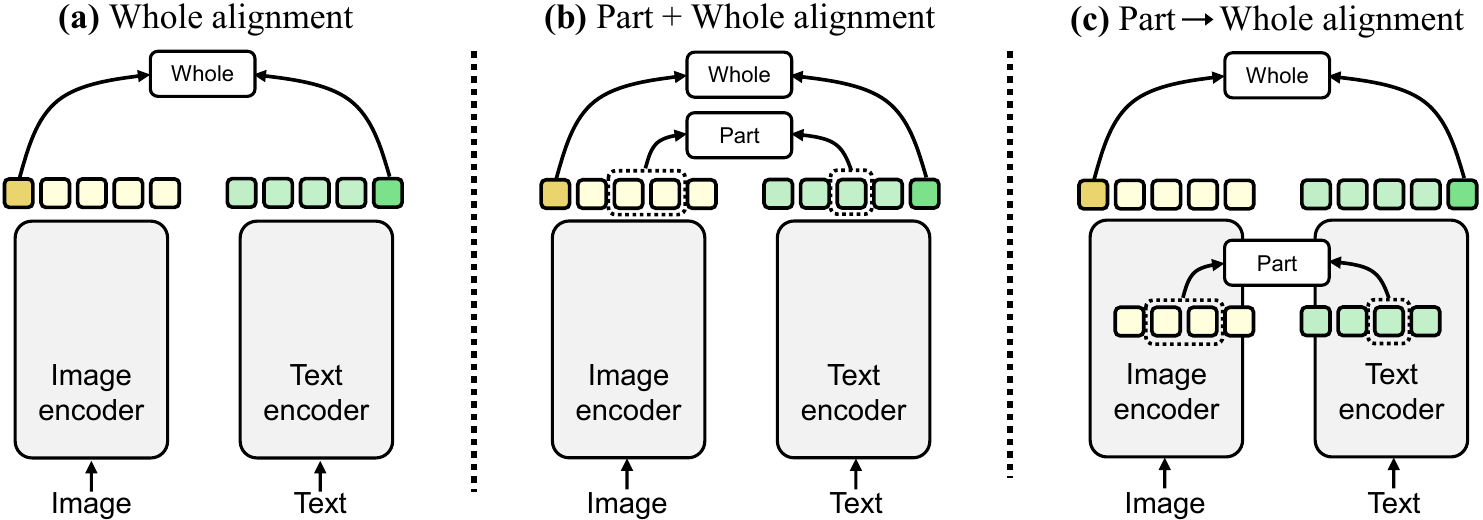}
    \par\vspace{0.5mm}
    \caption{%
        \textbf{Hierarchical \textit{Part $\rightarrow$ Whole} alignment is key to global scene understanding.}
        We compare three alignment strategies: (a) \textit{Whole}, (b) \textit{Part + Whole}, and (c) \textit{Part $\rightarrow$ Whole}.
        The table presents zero-shot long-caption retrieval performance (R@1) across five benchmarks, with the corresponding alignment strategies illustrated below.
        Our hierarchical \textit{Part $\rightarrow$ Whole} alignment substantially outperforms both alternative alignment strategies.
    }
    \label{fig:ablation_alignment}
\end{figure}

{\bf Setup.}
To examine whether hierarchical alignment offers an advantage over simply adding part-level alignment in a flat manner, we compare three alignment strategies while keeping the architecture fixed (\cref{fig:ablation_alignment}).
{\it Whole} applies only whole-level alignment at the final layer.
{\it Part + Whole} additionally introduces part-level alignment at the same final layer.
{\it Part $\rightarrow$ Whole} (Ours) establishes part-level alignment at intermediate layers and whole-level alignment at the final layer.

{\bf Results.}
As shown in \cref{fig:ablation_alignment}, adding part-level alignment at the final layer yields almost no improvement over whole-level alignment alone (72.6 vs. 72.5), suggesting that introducing local alignment alongside global alignment in a flat manner is insufficient to improve global understanding.
In contrast, our hierarchical {\it Part $\rightarrow$ Whole} strategy achieves a substantial {\bf +5.6} gain over whole-level alignment alone.
This contrast highlights that the benefit of part-level supervision depends not merely on introducing local alignment, but on how local and global alignment are organized within the representation hierarchy.
More broadly, these findings support our central hypothesis: whole scene understanding is more effectively learned when grounded part-level semantics are established at intermediate representations and subsequently composed into the final whole-level representation.


{\bf Comparison with FG-CLIP.}
A similar limitation of flat alignment is observed in FG-CLIP~\citep{xie2025fg}: adding region-level supervision improves local recognition but provides no meaningful gain in DCI long-caption retrieval (61.95 to 61.75 R@1).
Specifically, FG-CLIP combines global image-text alignment with bounding-box-supervised region-text alignment, applying both local and global objectives at the final layer.
This observation is consistent with our comparison between {\it Whole} and {\it Part + Whole}, suggesting that additional local supervision alone may not translate into better global understanding.
Extending our hierarchical alignment principle to models with explicit region-level annotations could therefore be a promising direction.

\subsection{Ablation on intermediate supervision}
\label{sec:ablation_auxiliary}

\ablationquestion{2}{What does grounding local semantics offer beyond intermediate supervision alone?}

{\bf Setup.}
To test whether the gains in whole-scene understanding arise from grounding local semantics rather than merely from deep supervision~\citep{lee2015deeply}, we compare our part-level alignment loss with an alternative intermediate supervision objective.
Specifically, rather than grounding sub-captions in intermediate visual segment representations as in CAFT, the alternative directly aligns each sub-caption embedding $t_{jk}^{\mathrm{part}}$ with the intermediate image-level [CLS] representation using a multi-positive sigmoid loss~\citep{xiao2025flair}.
We keep the supervision depth, loss weight, and final whole-level alignment objective fixed.

\noindent
\begin{minipage}[t]{0.55\linewidth}
    \vspace{0pt}
    {\bf Results.}
    As shown in \cref{tab:ablation_auxiliary}, image-level intermediate supervision provides only a modest gain over whole-level training (72.5 to 73.4), whereas text-grounded part alignment achieves {\bf 78.1}, a further {\bf +4.7} improvement.
    This gap suggests that the benefit of intermediate alignment cannot be explained by deep supervision alone.
    We argue that sub-caption-conditioned visual grounding encourages intermediate representations to capture localized semantics, which subsequent layers can integrate into whole-level understanding.
\end{minipage}
\hfill
\begin{minipage}[t]{0.43\linewidth}
    \vspace{0pt}
    \centering
    \captionof{table}{%
        \textbf{Intermediate supervision control.}
        Comparison of intermediate objectives with the final-layer alignment fixed.
    }
    \label{tab:ablation_auxiliary}
    \small
    \setlength{\tabcolsep}{3pt}
    \renewcommand{\arraystretch}{1.12}
    \begin{tabularx}{\linewidth}{@{}
    >{\hsize=.85\hsize\raggedright\arraybackslash}X
    >{\hsize=1.15\hsize\raggedright\arraybackslash}X
    c@{}}
    \toprule
    Final & Intermediate & Average R@1 \\
    \midrule
    Whole & None & 72.5 \\
    Whole & Image-level & 73.4 \\
    \rowcolor{gray!10}
    Whole & \textbf{Text-grounded} & \textbf{78.1} \\
    \bottomrule
\end{tabularx}
\end{minipage}

\begin{table}[!htbp]
    \centering
    \small
    \setlength{\tabcolsep}{5.0pt}
    \caption{%
        \textbf{Ablation study on components.}
        We ablate three main components of CAFT on CC3M-recap:
        visual hierarchy ({\bf VH}) replaces ViT with the fine-to-coarse image encoder, textual hierarchy ({\bf TH}) replaces the long-text encoder with the part-whole text encoder, and hierarchical alignment ({\bf HA}) adds additional part-level alignment on top of the whole-level alignment loss.
    }
    \vspace{5pt}
    \begin{tabular}{@{}c ccc c cc cc cc cc cc@{}}
        \toprule
        \multirow{2}{*}{ }
        & \multicolumn{3}{c}{Method}
        & \multirow{2}{*}{Average}
        & \multicolumn{2}{c}{DCI}
        & \multicolumn{2}{c}{DOCCI}
        & \multicolumn{2}{c}{SV-10k}
        & \multicolumn{2}{c}{Urban-1k}
        & \multicolumn{2}{c}{IIW} \\
        \cmidrule(lr){2-4}
        \cmidrule(lr){6-7}
        \cmidrule(lr){8-9}
        \cmidrule(lr){10-11}
        \cmidrule(lr){12-13}
        \cmidrule(lr){14-15}
        & {\bf VH} & {\bf TH} & {\bf HA}
        &
        & I2T & T2I
        & I2T & T2I
        & I2T & T2I
        & I2T & T2I
        & I2T & T2I \\
        \midrule
        (1) &  &  &
        & 69.5
        & 51.6 & 50.2
        & 59.4 & 56.5
        & 76.7 & 76.4
        & 74.2 & 76.7
        & 87.4 & 85.5 \\
        (2) & $\checkmark$ &  &
        & 71.2
        & 53.7 & 51.3
        & 62.2 & 58.5
        & 78.8 & 78.5
        & 76.2 & 76.9
        & 89.2 & 86.4 \\
        (3) &  & $\checkmark$ &
        & 68.7
        & 48.3 & 45.9
        & 58.7 & 55.1
        & 78.6 & 77.9
        & 76.3 & 77.2
        & 85.1 & 84.2 \\
        (4) &  & $\checkmark$ & $\checkmark$
        & 74.4
        & 53.4 & 54.4
        & 64.3 & 62.7
        & 83.0 & 81.8
        & 80.7 & 84.6
        & 90.0 & 89.2 \\
        \rowcolor{gray!10}
        (5) & $\checkmark$ & $\checkmark$ & $\checkmark$
        & \textbf{78.1}
        & \textbf{57.6} & \textbf{59.6}
        & \textbf{67.3} & \textbf{67.1}
        & \textbf{86.2} & \textbf{85.4}
        & \textbf{83.6} & \textbf{87.2}
        & \textbf{93.6} & \textbf{93.3} \\
        \bottomrule
    \end{tabular}
    \label{tab:ablation}
    \vspace{-10pt}
\end{table}

\subsection{Ablation study on main components}
\label{sec:ablation_components}

\ablationquestion{3}{How do our visual and textual encoders complement hierarchical alignment?}

{\bf Setup.}
To examine how effectively our visual and textual encoders support hierarchical vision-language learning principle, we evaluate different combinations of three components: visual hierarchy (VH), textual hierarchy (TH), and hierarchical alignment (HA).
We start from a LongSigLIP baseline, which extends the CLIP Text Transformer context length from 77 to 248~\citep{zhang2024long} and uses a whole-level SigLIP loss~\citep{zhai2023sigmoid}.
VH replaces ViT with the fine-to-coarse image encoder, TH replaces the long-text encoder with the part-whole text encoder, and HA replaces the whole-level SigLIP loss with our hierarchical part- and whole-level alignment loss.

{\bf Results.}
\cref{tab:ablation} shows the contribution of each component over the baseline.
From (1)$\rightarrow$(2), replacing ViT with our fine-to-coarse image encoder yields a moderate {\bf +1.7} gain, indicating that uncovering object-aware parts is beneficial even under global image-text alignment.
From (1)$\rightarrow$(3), replacing the long-text encoder with the part-whole text encoder causes a small {\bf -0.8} drop, suggesting that explicit part-text embeddings require alignment guidance to improve whole-text representations.
From (3)$\rightarrow$(4), adding hierarchical alignment gives the largest gain of {\bf +5.7}, supporting that whole-level understanding is strengthened when built from aligned part semantics.
From (4)$\rightarrow$(5), the fine-to-coarse image encoder becomes more effective when combined with hierarchical alignment, yielding a {\bf +3.7} gain compared to {\bf +1.7} without it.
%
Overall, combining all three components improves the baseline by {\bf +8.6}.
These results highlight the complementary roles of CAFT's hierarchical encoders and alignment objective, with hierarchical alignment playing a central role in connecting part-level semantics to whole-level understanding.

\section{Conclusion}
We introduce \sname, a hierarchical vision-language learning framework that aligns visual and textual structure across semantic scales.
Our results show that hierarchical cross-domain alignment is key to unsupervised visually grounded understanding from long captions, avoiding the failure modes of global-only or flat alignment. This improves both image-to-text and text-to-image retrieval at scale and yields more accurate zero-shot referring segmentation as a natural byproduct.



\newpage
\begin{ack}
This work was supported in part by the U.S. National Science Foundation (NSF) under Grant No. 2313151 to S. Yu, the Agency for Defense Development grant funded by the Korea government under No. 912A45701 to B. Woo and B. Pak, and the South Korean Institute of Information \& Communications Technology Planning \& Evaluation (IITP) under No. RS-2026-25524173 to S. Mo.

The research was initiated while B. Woo was at ADD and completed at the University of Michigan. We thank Junwha Hong for assistance with the distributed computing setup, Seokmin Lee and Sol Park for insightful feedback, and Hoseong Kim for constructive discussions and support.

The authors declare no competing interests.
\end{ack}


\bibliographystyle{plainnat}
\bibliography{reference}







\appendix





\label{suupl:full}
\onecolumn




\section*{Appendix}
In the supplemental material, we provide:
\vspace{0.5em}

\noindent
\hspace*{1.5em} \hyperref[supp:exp_setup]{A: Additional Experimental Details}

\hspace*{3.5em} \hyperref[supp:training_datasets]{A.1: Training Datasets} 

\hspace*{3.5em} \hyperref[sec:suppl_benchmark]{A.2: Evaluation Datasets} 

\hspace*{3.5em} \hyperref[supp:training_config]{A.3: Training Configuration} 

\hspace*{3.5em} \hyperref[supp:computation]{A.4: Computational Complexity} 

\hspace*{1.5em} \hyperref[supp:imple_details]{B: Additional Implementation Details}

\hspace*{3.5em} \hyperref[supp:superpixel]{B.1: Superpixel Tokenization}

\hspace*{3.5em} \hyperref[sec:chunking]{B.2: Chunking}

\hspace*{3.5em} \hyperref[supp:inference]{B.3: Inference}

\hspace*{1.5em} \hyperref[supp:exp]{C: Additional Experimental Results}

\hspace*{3.5em} \hyperref[supp:short_text]{C.1: Short-text Image Retrieval} 

\hspace*{3.5em} \hyperref[supp:vit_backbone]{C.2: ViT Backbone Control}

\hspace*{3.5em} \hyperref[supp:counterfactual]{C.3: Counterfactual Interventions}

\hspace*{3.5em} \hyperref[supp:chunking_sensitivity]{C.4: Caption Chunking Sensitivity}

\hspace*{3.5em} \hyperref[supp:fgovd]{C.5: Fine-grained Open-Vocabulary Object Detection}

\hspace*{3.5em} \hyperref[supp:data_overlap]{C.6: Training--Evaluation Data Overlap}

\hspace*{1.5em} \hyperref[supp:rel_work]{D: Additional Related Works}

\hspace*{1.5em} \hyperref[sec:suppl_more_vis]{E: Additional Visualization}

\hspace*{3.5em} \hyperref[supp:attn]{E.1: Attention Map Visualization}

\hspace*{3.5em} \hyperref[supp:multi-granuarity]{E.2: Attention Map Visualization across Multi-Granularity}






\newpage
\section{Additional Experimental Details}
\label{supp:exp_setup}
\subsection{Training Datasets}
\label{supp:training_datasets}
\sname is trained on DreamLIP~\citep{zheng2024dreamlip} datasets, which consist of CC3M-recap, CC12M-recap, YFCC15M-recap and Merged-30M. 
In these datasets, images are re-captioned with both short and long synthetic captions generated by a diverse set of MLLMs, and we jointly use multiple types of captions during training.
For the original CC3M, CC12M, and YFCC15M datasets, some images are no longer available due to their web-based nature; we therefore report the number of images actually used in our experiments for each dataset, along with the corresponding training GPU-hours, in \cref{tab:training_time_gpu}. All experiments were conducted using 8 A100 GPUs (80GB) and training on Merged-30M required approximately 5 days.
\vspace{-3mm}
\begin{table}[h]
    \centering
    \caption{Number of images and the corresponding GPU-hours for each dataset scales.}
    \label{tab:training_time_gpu}
    \begin{tabular}{lcccc}
        \toprule
        Dataset & \# Images & Training Hours & GPU Hours (Days) \\
        \midrule
        CC3M-recap  & \phantom{0}2.82M & \phantom{0}12.3  & \phantom{0}98.6 (\phantom{0}4.1)  \\
        CC12M-recap & 10.01M & \phantom{0}42.5   & 340.0 (14.2) \\
        YFCC15M-recap & 14.07M & \phantom{0}59.5   & 476.0 (19.8) \\
        Merged-30M & 26.90M & 114.4  & 915.2 (38.1) \\
        \bottomrule
    \end{tabular}
\end{table}

\subsection{Evaluation Datasets}
\label{sec:suppl_benchmark}

We present the detailed statistics of the six long-text-image retrieval datasets we use in \cref{tab:long_text_image_retrieval}. For DOCCI, we use only the test split for evaluation. 
For DCI, we prepend the standard caption to the long description, following \cite{xiao2025flair}.

\begin{table}[h]
\centering
\caption{Statistics of the long-text-image retrieval datasets.}
\begin{tabular}{lcccc}
\toprule
Dataset & \# Images & \# Texts & \# Sub-captions per Text & \# Tokens per Text \\
\midrule
DCI              & \phantom{0}7,805  & \phantom{0}7,805  & 10.81            & 172.73\\
DOCCI            & \phantom{0}5,000  & \phantom{0}5,000  & \phantom{0}7.12  & 141.52\\
ShareGPT4v-1k    & \phantom{0}1,000  & \phantom{0}1,000  & \phantom{0}8.15  & 173.24\\
ShareGPT4v-10k   & 10,000            & 10,000            & \phantom{0}8.24  & 173.66\\
Urban-1k         & \phantom{0}1,000  & \phantom{0}1,000  & \phantom{0}5.97  & 131.36\\
IIW              & \phantom{00}612   & \phantom{00}612   & 10.16            & 239.73 \\
\bottomrule
\end{tabular}
\label{tab:benchmark}
\end{table} 

\subsection{Training Configuration}
\label{supp:training_config}
We follow training configuration of \citep{xiao2025flair} as displayed in \cref{tab:training-config}. However, we use 2K batch size for all datasets, due to GPU RAM limit. Following SigLIP~\citep{zhai2023sigmoid}, the learnable temperature parameters ($\tau_{\text{p}}$, $\tau_{\text{w}}$) and bias terms ($b_{\text{p}}$, $b_{\text{w}}$) are initialized to 0.07 and -10, respectively. 

\begin{table}[h]
\centering
\caption{Training configuration for different datasets.}
\begin{tabular}{lcccc}
\toprule
Config & CC3M-recap & CC12M-recap & YFCC15M-recap & Merged-30M \\
\midrule
Batch size     & \multicolumn{4}{c}{2,048} \\
Optimizer      & \multicolumn{4}{c}{AdamW \citep{loshchilov2017decoupled}} \\
Learning rate  & \multicolumn{4}{c}{$5 \times 10^{-4}$} \\
Weight decay   & 0.5 & 0.5 & 0.5 & 0.2 \\
Adam $\beta$   & \multicolumn{4}{c}{$\beta_1, \beta_2 = (0.9, 0.98)$} \\
Adam $\epsilon$& $1 \times 10^{-8}$ & $1 \times 10^{-8}$ & $1 \times 10^{-8}$ & $1 \times 10^{-6}$ \\
Total epochs   & \multicolumn{4}{c}{32} \\
Warm up        & \multicolumn{4}{c}{2,000 (steps)} \\
LR scheduler   & \multicolumn{4}{c}{cosine decay} \\
\bottomrule
\end{tabular}
\label{tab:training-config}
\end{table}

\subsection{Computational Complexity}
\label{supp:computation}

We compare the computational cost of CAFT with two representative competitors:
FLAIR~\citep{xiao2025flair}, which represents short-text CLIP models, and LongCLIP~\citep{zhang2024long}, which represents
long-text CLIP models~(\cref{tab:complexity}). We report GFLOPs per
single image-text pair. For all models, the vision encoder FLOPs are measured
with a single $224 \times 224$ image input under a ViT-B-scale architecture.
For the text encoder, we use the corresponding caption setting of each model.

\begin{table}[h]
\centering
\caption{
 {\bf Computational complexity comparison.} We report GFLOPs per single image-text
pair, separately for the image encoder, text encoder, and their total cost.
CAFT is computationally cheaper overall than both FLAIR and LongCLIP.
}
\vspace{2mm}
\label{tab:complexity}
\begin{tabular}{lccc}
\toprule
Model & Image Encoder & Text Encoder & Total  \\
\midrule
FLAIR    & 18.46 &  {\bf 3.08} & 21.54 \\
LongCLIP & 18.32 & 10.91 & 29.23 \\
CAFT     & {\bf 11.34} &  8.27 & {\bf 19.61} \\
\bottomrule
\end{tabular}
\end{table}

{\bf Image encoder.}
While both LongCLIP and FLAIR use a standard ViT image encoder, CAFT replaces it
with the fine-to-coarse image encoder~(\cref{sec:part-whole-structure-in-vision}).
This design substantially reduces the computational cost of the image encoder
from about 18.3 GFLOPs to 11.3 GFLOPs. Specifically, the encoder progressively
reduces the number of visual tokens from 196 to 16 through fine-to-coarse token
grouping, which lowers the cost of later Transformer blocks. This indicates that
the fine-to-coarse image encoder is not introduced merely to scale up the model
for better performance; rather, it is an intentional architectural choice for
hierarchical visual representation learning, while also being computationally
more efficient.

{\bf Text encoder.}
FLAIR uses a standard CLIP-style Text Transformer with context length 77, whereas
LongCLIP extends the CLIP Text Transformer context length to 248, a common setting
in long-text CLIP variants~\citep{zhang2024long,najdenkoska2024tulip,
asokan2025finelip,xie2025fg,wu2024lotlip}. CAFT processes part texts using a
Part-text Transformer with context length 77, and then aggregates the resulting
part-text embeddings with a Whole-text Transformer with context length 4.
Although the part-whole text encoder requires more computation than FLAIR's text
encoder, which truncate captions longer than 77 tokens, it is more efficient
than standard long-context text encoders. This efficiency comes from the
quadratic cost of self-attention: instead of applying attention over a single
248-token sequence, CAFT follows a divide-and-conquer strategy that encodes
shorter part texts and then composes them at the whole-text level.

{\bf Total.}
As a result, CAFT requires 19.61 GFLOPs in total, which is lower than LongCLIP by
32.9\% and lower than FLAIR by 9.0\%. This shows that the architectural changes
in CAFT are motivated by the intended hierarchical representation design, rather
than by simply increasing computational scale, while still remaining
computationally efficient.

\newpage
\section{Additional Implementation Details}
\label{supp:imple_details}

\subsection{Superpixel Tokenization}
\label{supp:superpixel}

Instead of using square, uniform grid patch tokens as in ViT~\citep{dosovitskiy2020image}, we use superpixel tokens as visual units in our fine-to-coarse image encoder.
Given an input image, we first generate superpixels and then obtain each superpixel token by averaging the convolutional pixel representations within the corresponding superpixel region. 
To match the number of tokens produced by a ViT with $16 \times 16$ patches on a $224 \times 224$ image, we partition each image into 196 superpixels.

Superpixels are commonly generated using classical superpixel algorithms. 
For example, SuiT~\citep{lew2024superpixel} adopts SLIC~\citep{achanta2012slic} for superpixel tokenization, while CAST~\citep{ke2023cast} uses SEEDS~\citep{van2012seeds}. 
However, both SLIC and SEEDS rely on CPU-based iterative procedures, which prevent fully end-to-end GPU processing and can introduce a dataloader bottleneck during pretraining. 
This bottleneck becomes particularly problematic for language-image pretraining, where large batch sizes, e.g., 2K, are often required.
Moreover, these methods often struggle to capture accurate contours of thin objects. 
To address these limitations, we build our superpixel tokenization pipeline upon SFCN~\citep{yang2020superpixel}, a lightweight deep network for superpixel segmentation.
This design enables end-to-end GPU processing for superpixel token generation, making the tokenization process compatible with large-batch language-image pretraining while better preserving thin object boundaries, such as a broom (see \cref{fig:multi_granularity}).

\noindent\textbf{SFCN initialization.}
We use the released SFCN checkpoint pretrained on BSDS500 with region-reconstruction and spatial-compactness losses, and keep its weights frozen throughout CAFT training. The BSDS500 region annotations provide low-level boundary cues, but no semantic category labels.

\subsection{Chunking}
\label{sec:chunking}
In \cref{sec:part-whole-structure-in-language}, we briefly introduced the idea of dividing a long caption into $N$ sub-captions (chunks). This section provides a detailed description of the chunking strategies used in training and inference. Specifically, we adopt \textbf{random chunking} during training and \textbf{balanced chunking} during inference.  

Chunking begins by splitting a long caption into $L$ sentences:
\[
S_1, S_2, \ldots, S_L,
\]
where $L$ is a variable length depending on each sample.  

\textbf{Random chunking.}  
During training, each chunk is formed to contain 1, 2, or 3 sentences, with the number chosen randomly. We enforce two constraints: (1) each chunk must contain at least one sentence, and (2) no chunk may contain more than three sentences. Therefore, when $L > 3N$, excess sentences are discarded. Conversely, if $L < N$, sentences are randomly resampled with replacement until all chunks are filled. We use $N=4$; the average number of sub-captions per image for each dataset, corresponding to the mean of $L$, is reported in \cref{tab:benchmark}. 

\textbf{Balanced chunking.}  
During inference, results may vary depending on how sentences are distributed across chunks. To ensure stable and reproducible inference, we employ balanced chunking, which divides sentences as fairly as possible. Each chunk first receives $\lfloor L/N \rfloor$ sentences, and the remaining sentences are allocated to the earlier chunks in order. 

For example:
\[
\text{if } L = 6, N = 4 \;\;\Rightarrow\;\; [2,2,1,1],
\]
\[
\text{if } L = 11, N = 4 \;\;\Rightarrow\;\; [3,3,3,2].
\]

It is worth noting that our divide-and-conquer approach with chunking is also computationally efficient under the quadratic self-attention mechanism~\citep{song2024hierarchical}. In contrast, recent CLIP variants for long-text understanding~\citep{zhang2024long,wu2024lotlip,asokan2025finelip,najdenkoska2024tulip} process up to 248 tokens at once, leading to substantial computational cost.

Sensitivity analyses of the number of chunks, chunking strategy (random vs. balanced during inference), and sentence order are provided in \cref{supp:chunking_sensitivity}.

\newpage
\subsection{Inference}
\label{supp:inference}

\paragraph{Selection of inference weight $\alpha$.}
As described in \cref{sec:part-to-whole-vision-languae-alignment}, CAFT computes the image-text alignment score at inference time as a weighted sum of whole-level and part-level similarities, controlled by $\alpha$.
We select $\alpha=0.3$ based on retrieval R@1 on a randomly sampled 30K subset (approximately 1\%) of DreamLIP-3M training data, without using evaluation benchmarks.
As shown in \cref{tab:ablationPartScore}, this choice also achieves the best overall performance across four zero-shot benchmarks.
Notably, whole-only inference ($\alpha=0.0$) remains competitive, suggesting that the learned part-level semantics are reflected in the global representation, while incorporating part-level similarity further improves retrieval.
\vspace{-3mm}

\begin{table}[h]
\centering
\small
\setlength{\tabcolsep}{3pt}
\renewcommand{\arraystretch}{1.15}
\caption{%
    \textbf{Selection and ablation of the inference weight $\alpha$.}
    We select $\alpha$ using a randomly sampled 30K subset (approximately 1\%) of DreamLIP-3M training data, without using any evaluation benchmark.
    We additionally report zero-shot retrieval R@1 across four benchmarks to analyze sensitivity to $\alpha$.
    $\alpha=0.0$ and $\alpha=1.0$ correspond to whole-only and part-only inference, respectively.
}
\label{tab:ablationPartScore}
\vspace{3mm}

\begin{tabularx}{0.84\linewidth}{
    @{}c
    *{2}{>{\hsize=1.6\hsize\centering\arraybackslash}X}
    |
    *{8}{>{\hsize=0.85\hsize\centering\arraybackslash}X}
    @{}
}
\toprule
\multirow{3}{*}{$\alpha$}
& \multicolumn{2}{c|}{Selection}
& \multicolumn{8}{c}{Zero-shot Evaluation} \\
\cmidrule(lr){2-3}
\cmidrule(l){4-11}
& \multicolumn{2}{c|}{DreamLIP-3M (30K)}
& \multicolumn{2}{c}{DCI}
& \multicolumn{2}{c}{SV-10k}
& \multicolumn{2}{c}{Urban-1k}
& \multicolumn{2}{c}{IIW} \\
& I2T & T2I
& I2T & T2I
& I2T & T2I
& I2T & T2I
& I2T & T2I \\
\midrule
0.0
& 97.4 & 91.9
& 69.4 & 69.4
& 95.1 & 94.5
& \textbf{93.6} & 93.9
& 97.4 & 96.1 \\
\rowcolor{gray!10}
0.3
& \textbf{97.5} & \textbf{92.3}
& \textbf{69.7} & 72.4
& \textbf{95.5} & \textbf{94.8}
& \textbf{93.6} & 95.4
& \textbf{97.5} & 97.4 \\
0.7
& 97.0 & 92.1
& 63.2 & \textbf{74.1}
& 94.3 & 94.3
& 92.2 & \textbf{96.3}
& 97.4 & \textbf{97.9} \\
1.0
& 94.3 & 90.8
& 50.5 & 71.6
& 90.3 & 92.3
& 86.2 & 94.6
& 95.3 & 97.4 \\
\bottomrule
\end{tabularx}
\end{table}

\paragraph{Efficient inference via top-$k$ reranking.}
While incorporating part-level similarity improves retrieval accuracy, it requires text-conditioned visual features for each image-text pair, introducing additional inference cost.
We therefore adopt a two-stage strategy: rank all candidates using whole-level similarity ($\alpha=0$), then rerank the top-$k$ candidates with combined whole- and part-level similarities ($\alpha=0.3$).
Global image-text embeddings and intermediate visual tokens can be precomputed, so only text-conditioned attention pooling is required for the shortlisted candidates.

We evaluate the accuracy-efficiency trade-off across different values of $k$ in \cref{tab:topk_tradeoff}.
End-to-end latency and peak GPU memory are measured on SV-10k using a single A100 GPU, while Average R@1 is computed across DCI, DOCCI, SV-10k, Urban-1k, and IIW.

\begin{table}[h]
\centering
\caption{
    \textbf{Inference accuracy-efficiency trade-off.}
    We compare whole-only inference, top-$k$ reranking with $\alpha=0.3$, and full part-level scoring.
    Latency and peak GPU memory are measured on SV-10k using a single A100 GPU.
    Average R@1 is computed over five long-text retrieval benchmarks.
}
\label{tab:topk_tradeoff}
\vspace{2mm}
\begin{tabular}{lccc}
\toprule
Inference mode & Latency (s) & Peak memory (GB) & Average R@1 \\
\midrule
Whole-only ($k=0$) & \textbf{30.74} & \textbf{6.14} & 87.04 \\
\rowcolor{gray!10}
Top-$k$ ($k=16$) & 31.94 & 6.39 & 87.83 \\
Top-$k$ ($k=64$) & 32.70 & 6.93 & 87.79 \\
Top-$k$ ($k=128$) & 33.49 & 8.23 & 87.77 \\
Top-$k$ ($k=256$) & 34.58 & 15.27 & 87.78 \\
Full scoring & 38.53 & 19.77 & \textbf{87.88} \\
\bottomrule
\end{tabular}
\end{table}

Remarkably, reranking only the top 16 candidates achieves nearly the same accuracy as full scoring (87.83 vs. 87.88 R@1), improving upon whole-only inference by 0.79 points.
At the same time, it substantially reduces latency (31.94 vs. 38.53 s) and peak GPU memory (6.39 vs. 19.77 GB) compared to full scoring.
Increasing $k$ further provides little additional accuracy benefit while increasing computational cost.
These results demonstrate that CAFT can retain the benefits of part-level matching with minimal inference overhead, making it practical for larger-scale retrieval.

\clearpage

\section{Additional Experimental Results}
\label{supp:exp}

\subsection{Short-text Image Retrieval}
\label{supp:short_text}

\noindent
\begin{minipage}[t]{0.58\textwidth}
We also evaluate CAFT under standard short-text image-text retrieval settings to assess its effectiveness beyond long-caption scenarios.
Following prior works~\citep{lavoie2024modeling,xiao2025flair}, we conduct image-text retrieval experiments on the validation splits of MSCOCO~\citep{lin2014microsoft} and Flickr30k~\citep{plummer2015flickr30k}, whose images are annotated with five concise captions.
To deal with short caption, we treat each short caption as a {\it single chunk} and process it with the Part-text Transformer.
We consider two comparison groups: long-text CLIP models and standard CLIP models.

\medskip

{\bf Comparison with long-text CLIP models.}
It is generally observed that short- and long-text retrieval performance can exhibit a trade-off~\citep{lavoie2026clip,zhang2024long}, as training on full long captions tends to emphasize long-text understanding. We therefore first compare CAFT with existing long-text CLIP models.
As shown in Table~\ref{tab:short_retrieval_longclip}, CAFT outperforms these models by a large margin across both MSCOCO and Flickr30k. 
These results show that CAFT retains strong short-text retrieval performance while improving long-caption retrieval.

\medskip

{\bf Comparison with standard CLIP models.}
We next compare CAFT with standard CLIP models, which serve as strong baselines for conventional short-text retrieval. As shown in Table~\ref{tab:short_retrieval_standardclip}, CAFT clearly outperforms widely used models such as OpenCLIP and SigLIP. 
Compared with recent state-of-the-art models such as SigLIP 2 and FLAIR, CAFT remains competitive on short-caption benchmarks, while achieving substantially stronger performance on long-caption retrieval, as shown by the Urban-1k results.
\end{minipage}
\hfill
\begin{minipage}[t]{0.39\textwidth}
    \centering
    \vspace{0pt}
    \includegraphics[width=\linewidth]{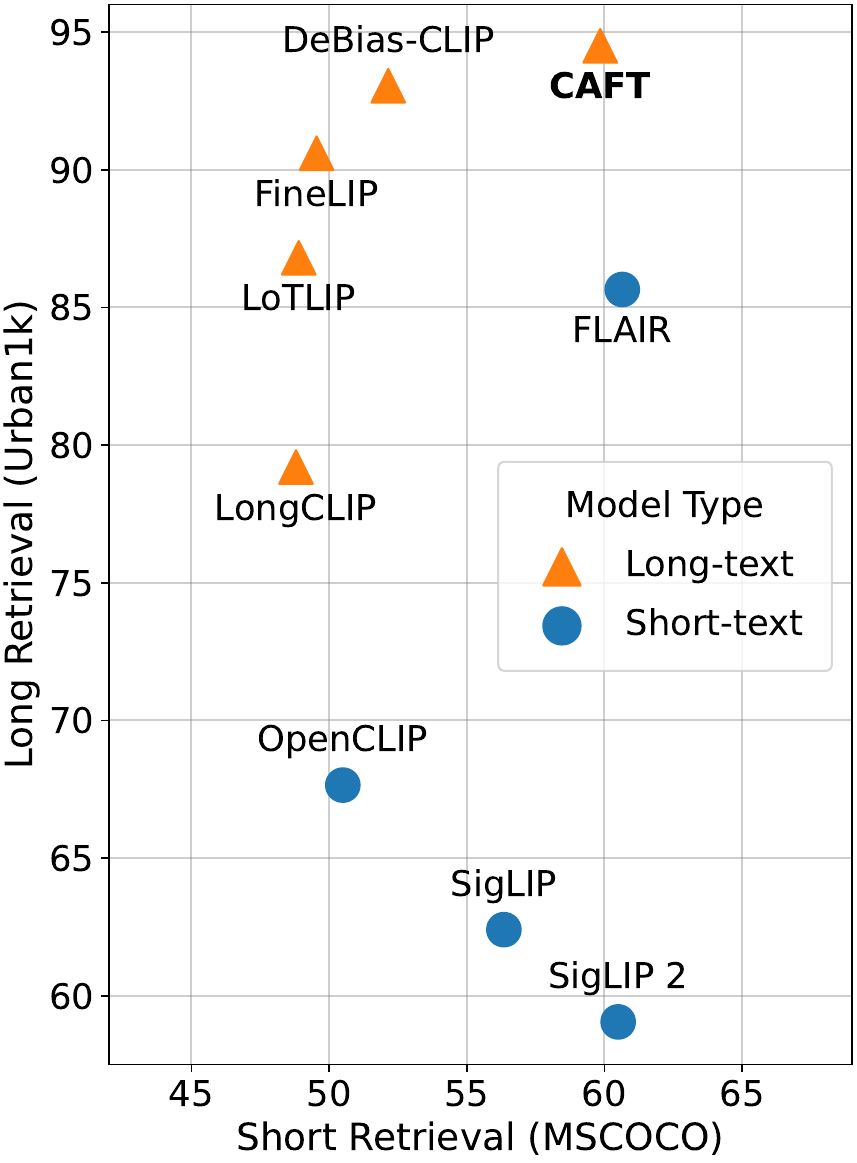}
    \captionof{figure}{{\bf Short- vs. long-text retrieval trade-off.}
    We plot short-text retrieval performance on MSCOCO along the horizontal axis and long-text retrieval performance on Urban1k along the vertical axis, where each score is averaged over image-to-text (I2T) and text-to-image (T2I) retrieval. CAFT lies in the upper-right region, indicating strong performance on both settings.}
    \label{fig:short_long_tradeoff}
\end{minipage}

We further visualize the short- and long-text retrieval trade-off across prior methods in \cref{fig:short_long_tradeoff}. CAFT lies in the upper-right region, indicating strong performance on both standard short-caption retrieval and long-caption retrieval benchmarks.

\begin{table}[h]
\centering
\caption{{\bf Comparison with long-text CLIP models on short-text retrieval.} 
CAFT substantially outperforms existing long-text CLIP models on MSCOCO and Flickr30k.}
\vspace{2mm}
\label{tab:short_retrieval_longclip}
\begin{tabular}{l | cc cc}
\toprule
\multirow{2}{*}{Method}
& \multicolumn{2}{c}{COCO}
& \multicolumn{2}{c}{Flickr30k} \\
& I2T & T2I & I2T & T2I \\
\midrule
LongCLIP~\citep{zhang2024long} & 57.3 & 40.3 & 85.9 & 70.7 \\
TULIP~\citep{najdenkoska2024tulip} & 56.8 & 40.7 & -- & -- \\
FineLIP~\citep{asokan2025finelip} & 58.7 & 40.4 & -- & -- \\
LoTLIP~\citep{wu2024lotlip} & 59.7 & 38.1 & 86.9 & 65.2 \\
DeBias-CLIP~\citep{lavoie2026clip} & 61.3 & 43.0 & -- & -- \\
\midrule
{\bf CAFT} (ours) & \bf 67.1 & \bf 52.6 & \bf 92.2 & \bf 81.0 \\
\bottomrule
\end{tabular}
\end{table}

\clearpage
\begin{table}[t]
\centering
\caption{{\bf Comparison with standard CLIP models on short-text retrieval.} CAFT outperforms SigLIP, and remains competitive with recent strong models on short-text retrieval, while achieving substantially stronger performance on long-caption retrieval (gray column).}
\vspace{2mm}
\label{tab:short_retrieval_standardclip}
\begin{tabular}{l | cc cc | aa}
\toprule
\multirow{2}{*}{Method}
& \multicolumn{2}{c}{COCO}
& \multicolumn{2}{c|}{Flickr30k}
& \multicolumn{2}{a}{Urban-1k} \\
& I2T & T2I & I2T & T2I & I2T & T2I \\
\midrule
OpenCLIP~\citep{pmlr-v139-radford21a} & 59.3 & 41.7 & 87.5 & 71.9 & 69.5 & 65.8 \\
MetaCLIP~\citep{xu2023demystifying} & 59.3 & 41.3 & 85.6 & 70.8 & 68.9 & 63.3 \\ 
SigLIP~\citep{zhai2023sigmoid} & 65.5 & 47.2 & 89.1 & 75.6 & 62.7 & 62.1 \\
SigLIP 2~\citep{tschannen2025siglip} & \bf 68.9 & 52.1 & 93.0 & 80.7 & 60.3 & 57.8 \\ 
FLAIR~\citep{xiao2025flair} & 68.0 & \bf 53.3 & \bf 94.7 & \bf 81.1 & 83.6 & 87.7 \\
\midrule
{\bf CAFT} (ours) & 67.1 & 52.6 & 92.2 & 81.0 & \bf 93.6 & \bf 95.4 \\
\bottomrule
\end{tabular}
\end{table}

\subsection{Ablation Study with a ViT Backbone}
\label{supp:vit_backbone}

We further investigate whether our hierarchical learning principle generalizes to a standard ViT backbone, beyond the proposed fine-to-coarse image encoder.
To this end, we repeat the alignment-strategy ablation in \cref{sec:ablation_strategies}, replacing our image encoder with a ViT while keeping the alignment objectives unchanged.
We compare whole-level alignment alone (\textit{Whole}), part- and whole-level alignment at the final layer (\textit{Part + Whole}), and part-level alignment at an intermediate layer followed by whole-level alignment at the final layer (\textit{Part $\rightarrow$ Whole}).
Specifically, part-level alignment is applied at the 6th or 11th Transformer layer, while whole-level alignment remains at the final layer.

\begin{table}[h]
\centering
\caption{
    \textbf{Alignment strategies with a ViT backbone.}
    We repeat the alignment-strategy ablation using a standard ViT with 196 visual tokens across all layers.
    Average R@1 is reported over five long-text retrieval benchmarks and all experiments are conducted on DreamLIP-3M.
}
\label{tab:vit_alignment}
\vspace{2mm}
\begin{tabular}{lccc}
\toprule
Method & Part-alignment layer & \# Visual tokens & Average R@1 \\
\midrule
Whole & -- & 196 & 68.7 \\
Part + Whole & Final (11th) & 196 & 68.6 \\
Part $\rightarrow$ Whole & Intermediate (6th) & 196 & \textbf{74.4} \\
\bottomrule
\end{tabular}
\end{table}

As shown in \cref{tab:vit_alignment}, intermediate part-level alignment outperforms both whole-only and final-layer part-level alignment by 5.7 and 5.8 R@1 points, respectively.
These results demonstrate that the benefit of hierarchical alignment extends to a standard ViT backbone, supporting the generality of our part-to-whole learning principle beyond the proposed fine-to-coarse architecture.

\subsection{Counterfactual Interventions}
\label{supp:counterfactual}

We investigate whether CAFT's global image-text similarity relies on correctly grounded local evidence through two complementary counterfactual interventions: visual region masking and textual phrase replacement.
We use 2,455 samples from the RefCOCOg subset of GranD-f~\citep{rasheed2024glamm}, which provides long captions with phrase-level segmentation annotations.

We measure the normalized drop in global image-text cosine similarity:
\begin{equation}
\Delta_i =
\frac{S(I_i,T_i)-S(I_i',T_i')}{\bar{S}},
\qquad
\bar{S} = \frac{1}{N}\sum_{i=1}^{N}S(I_i,T_i),
\end{equation}
where $(I_i',T_i')$ denotes the modified image-text pair and $\bar{S}$ is the mean original similarity for each model.
We report the mean normalized drop (\%) and compare CAFT against each baseline using paired one-sided Wilcoxon signed-rank tests.

\paragraph{Visual intervention.}
To examine whether global matching depends on local visual evidence, we randomly select an annotated phrase-mask pair for each image and gray out the corresponding region while keeping the caption unchanged.
As shown in \cref{tab:counterfactual_visual}, CAFT exhibits a substantially larger similarity drop (6.54\%) than LongCLIP (1.41\%) and DeBias-CLIP (3.02\%), with statistically significant differences ($p<0.001$).
This indicates that CAFT's global matching is more sensitive to the removal of relevant local visual evidence.

\begin{table}[h]
\centering
\caption{
    \textbf{Counterfactual visual intervention.}
    Mean normalized similarity drop (\%) after masking an annotated image region.
    The $p$-values are from paired one-sided Wilcoxon signed-rank tests against CAFT.
}
\label{tab:counterfactual_visual}
\vspace{2mm}
\begin{tabular}{lcc}
\toprule
Model & Similarity drop (\%) & $p$-value \\
\midrule
LongCLIP & 1.41 & $1.4 \times 10^{-53}$ \\
DeBias-CLIP & 3.02 & $9.8 \times 10^{-39}$ \\
CAFT & \textbf{6.54} & -- \\
\bottomrule
\end{tabular}
\end{table}

\paragraph{Textual intervention.}
We further examine whether global matching is sensitive to changes in local textual semantics by replacing an annotated entity or attribute phrase while keeping the image unchanged.
We consider two settings:
(1) \textit{Different-caption replacement}, which substitutes a phrase from another caption, introducing incorrect local evidence;
and (2) \textit{Same-caption replacement}, which substitutes another annotated phrase from the same caption.
The latter is applied only to captions containing at least two annotated phrases.
Since the replacement phrase still describes content present in the image, this setting provides a more challenging test of compositional understanding than introducing unrelated content.

\cref{tab:counterfactual_textual} shows that CAFT exhibits the largest similarity drop in both settings, with statistically significant differences against both baselines ($p<10^{-25}$).
Notably, CAFT remains more sensitive even under same-caption replacement (5.76\%), where the substituted phrase still refers to valid scene content.
Together with visual region masking, these results provide complementary evidence that CAFT's global matching depends more strongly on correctly grounded local semantics.

\begin{table}[h]
\centering
\caption{
    \textbf{Counterfactual textual interventions.}
    Mean normalized similarity drop (\%) when an annotated entity or attribute phrase is replaced with a phrase from a different or the same caption.
}
\label{tab:counterfactual_textual}
\vspace{2mm}
\begin{tabular}{lcc}
\toprule
Model & Different-caption & Same-caption \\
\midrule
LongCLIP & 3.66 & 1.06 \\
DeBias-CLIP & 10.11 & 2.91 \\
CAFT & \textbf{17.79} & \textbf{5.76} \\
\bottomrule
\end{tabular}
\end{table}

\subsection{Caption Chunking Sensitivity}
\label{supp:chunking_sensitivity}

\paragraph{Sentences per chunk.}
We first determine how many sentences to include in each chunk to avoid excessive token truncation.
On DreamLIP-3M, the percentages of chunks exceeding the 77-token context length are 0.02\%, 0.17\%, 12.11\%, and 70.82\% for 1, 2, 3, and 4 sentences, respectively.
Given the sharp increase at four sentences, we limit each chunk to 1--3 sentences.

\paragraph{Number of chunks.}
We next determine the number of chunks $N$ by balancing sentence resampling and discarding.
For a caption with $L$ sentences, $L<N$ requires resampling with replacement, while $L>3N$ leads to discarding excess sentences.
As shown in \cref{tab:chunk_processing}, we select $N=4$ as a trade-off: it substantially reduces sentence discarding compared with $N=3$ (2.49\% vs. 15.44\%), while avoiding the high resampling rate of $N=5$ (35.90\%).

\begin{table}[!htbp]
\centering
\small
\renewcommand{\arraystretch}{1.12}
\caption{
    \textbf{Caption processing rates on DreamLIP-3M.}
    Percentage of captions involving sentence resampling, sentence discarding, or token truncation for different numbers of chunks $N$.
}
\label{tab:chunk_processing}
\vspace{2mm}
\begin{tabular}{@{}cccc@{}}
\toprule
$N$ & Resampled (\%) & Discarded (\%) & Truncated (\%) \\
\midrule
3 & 0.00 & 15.44 & 7.25 \\
4 & 12.64 & 2.49 & 2.32 \\
5 & 35.90 & 0.15 & 0.97 \\
\bottomrule
\end{tabular}
\end{table}

\paragraph{Retrieval sensitivity to $N$.}
We further investigate how the number of chunks $N$ affects retrieval performance.
We train CAFT with $N\in\{3,4,5\}$ on DreamLIP-3M for four epochs, keeping all other settings identical.
As shown in \cref{tab:chunk_count_retrieval}, average R@1 increases from 62.0 to 62.7 and 63.8 as $N$ increases.
While $N=5$ achieves the highest accuracy in this four-epoch experiment, we retain $N=4$ based on the balance between sentence resampling and discarding.

\begin{table}[!htbp]
\centering
\small
\setlength{\tabcolsep}{4pt}
\renewcommand{\arraystretch}{1.12}
\caption{
    \textbf{Retrieval sensitivity to the number of chunks.}
    All variants are trained for four epochs on DreamLIP-3M.
    We report average R@1 and I2T / T2I R@1 on five benchmarks.
}
\label{tab:chunk_count_retrieval}
\vspace{2mm}
\begin{tabular}{@{}cc cc cc cc cc cc@{}}
\toprule
\multirow{2}{*}{$N$}
& \multirow{2}{*}{Average}
& \multicolumn{2}{c}{DCI}
& \multicolumn{2}{c}{DOCCI}
& \multicolumn{2}{c}{SV-10k}
& \multicolumn{2}{c}{Urban-1k}
& \multicolumn{2}{c}{IIW} \\
\cmidrule(lr){3-4}
\cmidrule(lr){5-6}
\cmidrule(lr){7-8}
\cmidrule(lr){9-10}
\cmidrule(l){11-12}
& & I2T & T2I & I2T & T2I & I2T & T2I & I2T & T2I & I2T & T2I \\
\midrule
3 & 62.0
& 42.9 & 42.6
& 51.0 & 47.9
& 71.0 & 69.9
& 64.2 & 69.3
& 80.9 & 80.1 \\
4 & 62.7
& 43.2 & 42.1
& 51.7 & 48.3
& 71.2 & 69.9
& 65.0 & 69.1
& 84.6 & \textbf{81.5} \\
5 & \textbf{63.8}
& \textbf{44.6} & \textbf{44.9}
& \textbf{52.5} & \textbf{49.7}
& \textbf{71.3} & \textbf{70.3}
& \textbf{66.7} & \textbf{70.7}
& \textbf{85.3} & \textbf{81.5} \\
\bottomrule
\end{tabular}
\end{table}

\paragraph{Fixed versus variable-length chunking.}
We further investigate whether grouping consecutive sentences into chunks is more effective than encoding each sentence separately.
We compare our default strategy, which groups 1--3 consecutive sentences into $N=4$ chunks, against a variable-length alternative where each sentence forms a separate chunk, padded to a maximum of 12 chunk tokens.
Both variants are evaluated under the same DreamLIP-3M setting.

As shown in \cref{tab:chunk_strategy}, our default strategy improves average R@1 by 1.5 points (78.1 vs. 76.6), outperforming the alternative across all ten retrieval metrics.
This suggests that grouping neighboring sentences provides useful local context beyond encoding each sentence independently.

\begin{table}[!htbp]
\centering
\small
\setlength{\tabcolsep}{4pt}
\renewcommand{\arraystretch}{1.12}
\caption{
    \textbf{Chunking strategy ablation on DreamLIP-3M.}
    We compare fixed 1--3-sentence grouping with variable-length single-sentence chunking.
    We report average R@1 and I2T / T2I R@1 across five benchmarks.
}
\label{tab:chunk_strategy}
\vspace{2mm}
\begin{tabular}{@{}l c cc cc cc cc cc@{}}
\toprule
\multirow{2}{*}{Strategy}
& \multirow{2}{*}{Average}
& \multicolumn{2}{c}{DCI}
& \multicolumn{2}{c}{DOCCI}
& \multicolumn{2}{c}{SV-10k}
& \multicolumn{2}{c}{Urban-1k}
& \multicolumn{2}{c}{IIW} \\
\cmidrule(lr){3-4}
\cmidrule(lr){5-6}
\cmidrule(lr){7-8}
\cmidrule(lr){9-10}
\cmidrule(l){11-12}
& & I2T & T2I & I2T & T2I & I2T & T2I & I2T & T2I & I2T & T2I \\
\midrule
1--3 sentences, $N=4$
& \textbf{78.1}
& \textbf{57.6} & \textbf{59.6}
& \textbf{67.3} & \textbf{67.1}
& \textbf{86.2} & \textbf{85.4}
& \textbf{83.6} & \textbf{87.2}
& \textbf{93.6} & \textbf{93.3} \\
Single sentence, $N\leq12$
& 76.6
& 56.3 & 57.5
& 66.5 & 65.3
& 85.0 & 84.6
& 81.9 & 85.2
& 92.5 & 90.9 \\
\bottomrule
\end{tabular}
\end{table}

\paragraph{Inference-time chunking strategy.}
We examine whether retrieval performance is sensitive to inference-time chunking, comparing balanced chunking with random chunking over five seeds using the same CAFT-30M checkpoint.
Balanced chunking achieves 87.9 average R@1, compared with $87.8\pm0.1$ for random chunking across five benchmarks.
The comparable performance indicates robustness to sentence grouping, while balanced chunking ensures deterministic inference.

\noindent
\begin{minipage}[t]{0.53\linewidth}
\vspace{0pt}
\paragraph{Sensitivity to sentence order.}
We further investigate whether retrieval depends on sentence order by randomly shuffling complete sentences at inference time.
This preserves individual sentence semantics while potentially altering cross-sentence references.
As shown in \cref{tab:sentence_order}, CAFT exhibits a substantially smaller performance drop (1.4 points) than LongCLIP (6.9) and FLAIR (8.0).
This suggests that independently encoding caption parts reduces reliance on sentence position and salient early-caption cues.
\end{minipage}
\hfill
\begin{minipage}[t]{0.44\linewidth}
\vspace{0pt}
\centering
\captionof{table}{
    \textbf{Sensitivity to sentence order.}
    Average R@1 across five benchmarks before and after sentence shuffling.
}
\label{tab:sentence_order}
\vspace{2mm}
\small
\setlength{\tabcolsep}{3pt}
\renewcommand{\arraystretch}{1.12}
\begin{tabular}{@{}lccc@{}}
\toprule
Model & Original & Shuffled & Change \\
\midrule
LongCLIP & 72.3 & 65.4 & $-6.9$ \\
FLAIR & 80.4 & 72.4 & $-8.0$ \\
CAFT & \textbf{87.9} & \textbf{86.4} & $\mathbf{-1.4}$ \\
\bottomrule
\end{tabular}
\end{minipage}

\subsection{Fine-Grained Open-Vocabulary object Detection (FG-OVD)}
\label{supp:fgovd}
We further evaluate the visual grounding capability of \sname on the Fine-Grained Open-Vocabulary object Detection (FG-OVD) benchmark~\citep{bianchi2024devil}. FG-OVD assesses a model's ability to match bounding box regions with fine-grained descriptions by asking it to distinguish the correct description from ten hard negatives.
\cref{tab:fg-ovd} demonstrates that \sname exhibits superior fine-grained understanding compared to widely used Vision-Language Models, such as CLIP~\citep{pmlr-v139-radford21a} and EVA-CLIP~\citep{sun2023eva}. Notably, while models trained on long captions often suffer from a lack of regional understanding~\citep{wu2024lotlip,bianchi2024devil}, \sname achieves a substantial performance margin over Long-CLIP~\citep{zhang2024long}. We attribute this improvement to \sname's robust visual grounding capability. For a fair comparison, we report results for FG-CLIP~\citep{xie2025fg} and FineCLIP~\citep{jing2024fineclip} separately, as these methods leverage dense bounding-box annotations during training. 

\begin{table}[h]
    \centering
    \caption{FG-OVD results across difficulty levels.}
    \label{tab:fg-ovd}
    \begin{tabular}{lcccc}
        \toprule
        Method & Hard & Medium & Easy & Trivial \\
        \midrule
        \multicolumn{5}{l}{\textit{Trained w/o dense annotations}} \\
        CLIP      & 12.0 & 23.1 & 22.2 & 58.5 \\
        EVA-CLIP  & 14.0 & 30.1 & 29.4 & 58.3 \\
        Long-CLIP & 9.2  & 18.4 & 16.2 & 51.8 \\
        \textbf{\sname} & \textbf{16.8} & \textbf{32.8} & \textbf{38.6} & \textbf{62.5} \\
        \midrule
        \multicolumn{5}{l}{\textit{Trained w/ dense annotations}} \\
        FineCLIP & 26.8 & 49.8 & 50.4 & 71.9 \\
        FG-CLIP  & 46.1 & 66.6 & 68.7 & 83.4 \\
        \bottomrule
    \end{tabular}
\end{table}

\subsection{Training--Evaluation Data Overlap}
\label{supp:data_overlap}

For fair comparison with prior work, we follow the established training and evaluation protocol used by FLAIR.
Our training data are drawn from CC3M, CC12M, and YFCC15M, widely used public web-sourced datasets for vision-language pretraining.
However, their web-based nature introduces the possibility of unintended overlap with evaluation images.
We therefore conduct an explicit overlap audit to assess its potential impact on our results.

Following DataComp~\citep{gadre2023datacomp}, we use the ISC21-winning copy-detection model~\citep{yokoo2021contrastive} with a similarity threshold of 0.604169.
Among 1,195 candidate pairs, manual inspection confirms 15 overlapping images out of 25,417 evaluation images (0.06\%): 7 in DCI, 6 in SV-10k, and 2 in Urban-1k.
We disclose all identified evaluation images in \cref{tab:overlap_images}.

To assess the impact of these overlaps, we remove the corresponding images and captions and reevaluate CAFT, LongCLIP, and DeBias-CLIP.
As shown in \cref{tab:overlap_retrieval}, the average R@1 changes by at most 0.02 points for every model.
These results confirm that the identified overlaps have negligible impact on retrieval performance and do not affect our conclusions.

\begin{table}[!htbp]
\centering
\small
\setlength{\tabcolsep}{6pt}
\renewcommand{\arraystretch}{1.12}
\caption{
    \textbf{Retrieval performance before and after overlap removal.}
    R@1 before $\rightarrow$ after removing overlapping evaluation images.
    Mean averages the six retrieval directions shown.
}
\label{tab:overlap_retrieval}
\vspace{2mm}
\begin{tabular}{@{}llccc@{}}
\toprule
Dataset & Direction & CAFT & LongCLIP & DeBias-CLIP \\
\midrule
\multicolumn{2}{l}{Mean}
& $86.89 \to 86.91$
& $70.48 \to 70.49$
& $84.15 \to 84.15$ \\
\midrule
\multirow{2}{*}{DCI}
& I2T & $69.68 \to 69.72$ & $57.66 \to 57.66$ & $67.61 \to 67.63$ \\
& T2I & $72.42 \to 72.50$ & $51.10 \to 51.10$ & $68.96 \to 68.97$ \\
\addlinespace[3pt]
\multirow{2}{*}{SV-10k}
& I2T & $95.47 \to 95.47$ & $76.59 \to 76.60$ & $90.30 \to 90.29$ \\
& T2I & $94.79 \to 94.81$ & $78.64 \to 78.64$ & $91.72 \to 91.71$ \\
\addlinespace[3pt]
\multirow{2}{*}{Urban-1k}
& I2T & $93.60 \to 93.59$ & $79.10 \to 79.16$ & $92.70 \to 92.69$ \\
& T2I & $95.40 \to 95.39$ & $79.80 \to 79.76$ & $93.60 \to 93.59$ \\
\bottomrule
\end{tabular}
\end{table}

\begin{table}[!htbp]
\centering
\small
\setlength{\tabcolsep}{12pt}
\renewcommand{\arraystretch}{1.12}
\caption{
    \textbf{Confirmed training--evaluation image overlaps.}
    All 15 overlapping evaluation images are listed by dataset.
    DCI filenames are relative to \texttt{dci/densely\_captioned\_images/photos/}.
}
\label{tab:overlap_images}
\vspace{2mm}
\begin{tabular}{@{}lll@{}}
\toprule
DCI (7) & SV-10k (6) & Urban-1k (2) \\
\midrule
\texttt{sa\_1552159.jpg} & \texttt{sa\_000007/images/sa\_85112.jpg} & \texttt{Urban1k/images/542.jpg} \\
\texttt{sa\_1546970.jpg} & \texttt{sa\_000040/images/sa\_452601.jpg} & \texttt{Urban1k/images/579.jpg} \\
\texttt{sa\_1549938.jpg} & \texttt{sa\_000002/images/sa\_26529.jpg} & \\
\texttt{sa\_1553651.jpg} & \texttt{sa\_000004/images/sa\_46609.jpg} & \\
\texttt{sa\_1547809.jpg} & \texttt{sa\_000007/images/sa\_87906.jpg} & \\
\texttt{sa\_1545429.jpg} & \texttt{sa\_000008/images/sa\_93454.jpg} & \\
\texttt{sa\_1553857.jpg} & & \\
\bottomrule
\end{tabular}
\end{table}

\newpage
\section{Additional Related Work}
\label{supp:rel_work}


\textbf{Hierarchical Attention Transformers} have been extensively explored in NLP domain to address the challenges of processing lengthy texts, particularly for tasks such as long document classification.
Drawing inspiration from early work HAN~\citep{yang2016hierarchical}, prior studies~\citep{wu2021hi,chalkidis2022exploration,he2024hdt} typically encode documents in a hierarchical fashion--encoding local segments individually and then aggregating them into document-level representation.
However, to the best of our knowledge, such hierarchical encoding strategies remain unexplored in the domain of language-image pretraining.
Inspired by the divide-and-conquer strategy with chunking~\citep{chalkidis2022exploration,song2024hierarchical}, our hierarchical text transformer processes distinct sub-caption embeddings and aggregates them into a holistic caption representation.
As this structured language modeling aligns well with the form of visual organization, it effectively prevents specific salient tokens from dominating the long caption~\citep{wu2024lotlip}.

\textbf{Textual hierarchies for vision-language alignment} are often inherited from linguistic structures such as syntax parses or semantic graphs. 
PowerCLIP~\citep{kawamura2025powerclip} employ a syntactic parser to align image regions with a textual parse tree. 
Structure-CLIP~\citep{huang2024structure} incorporates external semantic graph knowledge, encouraging structured image-text alignment.
Although these hierarchies help compositional understanding~\citep{yuksekgonul2022and}, they do not explicitly reflect how visual scenes are organized, lacking natural correspondence with the visual hierarchy.
In contrast, we focus on the spatially oriented textual hierarchy inherent in long captions, where each sentence can be associated with a corresponding visual component~\citep{zheng2024dreamlip}.

\textbf{Hierarchical vision-language models} have been relatively underexplored due to the difficulty of jointly aligning visual and textual structures.
HiCLIP~\citep{geng2023hiclip} employs a hierarchical architecture for vision and text, uncovering part-whole hierarchy for fine-grained recognition.
However, HiCLIP enforces alignment only at the global level, preventing the model from learning associations among local elements.
Both HiVLP~\citep{chen2022hivlp} and HieCLIP~\citep{hua2025hieclip} introduce multi-level alignment across different encoder depths, yet their representations remain unstructured.
In contrast, our approach takes advantage of both explicit hierarchical architectures and a hierarchical alignment loss, enabling structured vision-language understanding across both local and global semantics.

\figImTextHier{h}

\newpage
\section{Additional Visualization}
\label{sec:suppl_more_vis}

\subsection{Attention Map Visualization}
\label{supp:attn}
\begin{figure*}[h]
\centering
\small
    

\includegraphics[width=\linewidth]{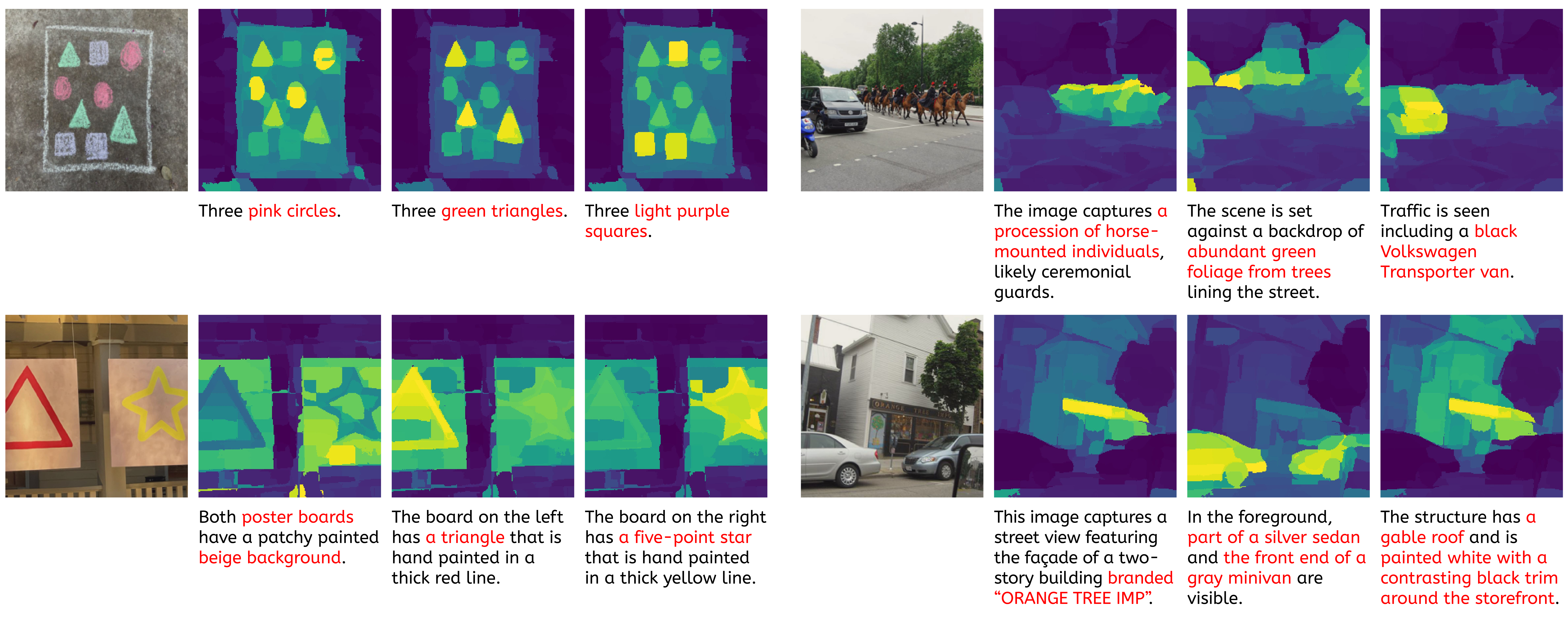}
\caption{
    \textbf{\sname provides spatially grounded, compositional visual grounding.}
    We visualize attention maps from the $\mathrm{AttnPool}$ module for three subcaptions in each example. The leftmost image is the input; the three heatmaps to the right show head-averaged attention with respect to each sub-caption. The scribble example shows \sname distinguishing \textit{pink circles} from \textit{green triangles}. Querying with the text \textit{ORANGE TREE IMP}, \sname grounds it accurately. These are preliminary examples, with possible confounding factors.
}
\label{fig:suppl_text_attn_map}
\end{figure*}

We present additional attention maps from the $\mathrm{AttnPool}$ module for each subcaption in \cref{fig:suppl_text_attn_map}. The visualizations illustrate how \sname grounds colors, shapes, and text in images. 

\begin{figure}[h]
    \centering
    \includegraphics[width=0.65\textwidth]{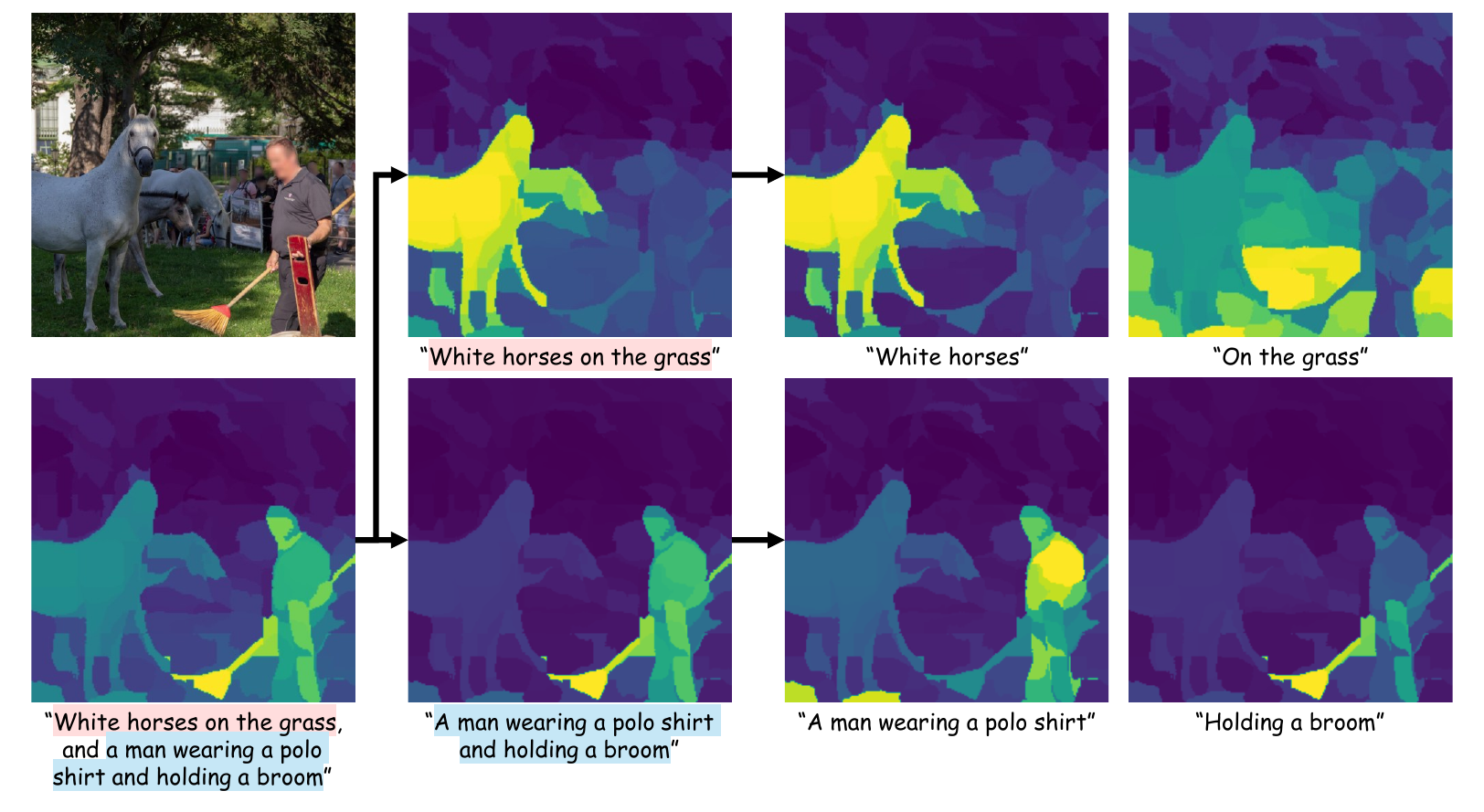}
    \caption{
        \textbf{\sname is capable of visual grounding across multiple levels of granularity}. A single caption may describe different objects simultaneously, and can be decomposed into smaller units. To reflect this hierarchical nature of caption, we decompose the sentence-level caption into smaller phrases and visualize their corresponding grounding results. Remarkably, \sname (1) can locate multiple objects at the same time (e.g., both \textit{a man} and \textit{horses} in the bottom left) and (2) precisely identify individual parts in the phrases (e.g., \textit{holding a broom} in the bottom right). 
    }
    \label{fig:multi_granularity}
\end{figure}

\subsection{Attention Map Visualization across Multi-Granularity}
\label{supp:multi-granuarity}
Although our training objective primarily targets the two-level spatial hierarchy that lies between the sub-caption and the full caption, our design philosophy is not restricted by the depth of hierarchy and can be naturally extended to deeper levels. We view \sname as a first step toward even more fine-grained alignment beyond the sentence-level (i.e., aligning forest, trees, and branches).
Nonetheless, we show that \sname effectively captures multi-granular semantics within sentence-level captions, despite not being explicitly trained for this, as shown in \cref{fig:multi_granularity}.
\sname not only localizes each object individually, but can also localize multiple objects simultaneously (e.g., both the \textit{man} and the \textit{horses}). Moreover, when we decompose a phrase (e.g., \textit{a man wearing a polo shirt and holding a broom}) into smaller parts (e.g., \textit{holding a broom}), \sname successfully grounds the corresponding part within the image.

\section{Limitations}
\label{supp:limitations}
\textbf{Synthetic data reliance.}
Similar to other long-text CLIP models, CAFT relies on synthetic long-caption datasets due to the lack of large-scale human-annotated images with lengthy and detailed captions.
In our experiments, we use the DreamLIP dataset, whose captions are generated by multimodal large language models (MLLMs).
These synthetic captions may contain unintended biases or noisy descriptions caused by hallucinations. 
Future work could study training with higher-quality human-verified long captions or more robust filtering of synthetic annotations.

\textbf{Fixed chunking of long captions.}
CAFT processes long captions by splitting them into a fixed number of part-text chunks.
Although we group captions at the sentence level when possible, each part-text may still mix multiple entities, or semantically connected descriptions may be separated across chunks.
Future work could explore more flexible caption decomposition strategies or encourage part-level distinctions within transformers with longer context lengths.

\section{Broader Impact}
\label{supp:broader_impact}

This work aims to improve fine-grained vision-language understanding by enabling models to ground detailed textual descriptions in corresponding image regions and compose them into whole-scene representations. 
Its potential positive impacts include more reliable image-text retrieval, improved accessibility tools that describe visual content in greater detail, and stronger vision-language systems for applications requiring faithful grounding, such as assistive technologies, educational tools, and visual search. 
At the same time, improved fine-grained image-text understanding could be misused in systems for surveillance, identity profiling, or large-scale visual content analysis without appropriate consent. 
The model may also inherit biases from large-scale image-text data, which can lead to uneven performance across demographic groups, cultures, or visual domains.

\section{Model Release and Safeguards}
\label{supp:safe}
\label{supp:release}

We have publicly released the CAFT model, source code, and documentation to support reproducibility and future research.
Although CAFT is not a generative model, it may be used in downstream applications involving image-text retrieval, ranking, or classification.
The release includes documentation on intended use, limitations, potential failure modes, and responsible-use guidelines.
We discourage use in privacy-sensitive, surveillance-oriented, biometric identification, or safety-critical applications without additional validation.
The released resources are available at \url{https://byeongju.me/CAFT/}.

\section{Existing Assets and Licenses}
\label{supp:licenses}
We use several existing datasets, models, and code assets for training, evaluation, and comparison. We cite the original papers or repositories for all such assets and use them only for research purposes, following their respective licenses and terms of use.

\begin{itemize}[leftmargin=1.2em, itemsep=0.2em, topsep=0.2em]
\item \textbf{DreamLIP re-captioned datasets.} We use the DreamLIP re-captioned datasets, including CC3M-recap, CC12M-recap, YFCC15M-recap, and Merged-30M, as training data. The project is available at \url{https://github.com/ant-research/DreamLIP}
 and is released under CC-BY-4.0. We follow the DreamLIP license and the terms of the original underlying datasets.
 
\item \textbf{CC3M / Conceptual Captions.} CC3M is used as one of the original data sources for CC3M-recap. The dataset is available at \url{https://github.com/google-research-datasets/conceptual-captions}. 
The dataset can be freely used for any purpose; acknowledgement of Google LLC as the data source is appreciated. The dataset is provided ``AS IS'' without warranty.

\item \textbf{CC12M / Conceptual 12M.} CC12M is used as one of the original data sources for CC12M-recap. The dataset is available at \url{https://github.com/google-research-datasets/conceptual-12m}
. The dataset can be freely used for any purpose; acknowledgement of Google LLC as the data source is appreciated. The dataset is provided ``AS IS'' without warranty, and Google disclaims liability for any direct or indirect damages resulting from its use.

\item \textbf{YFCC15M / YFCC100M subset.} YFCC15M is used as one of the original data sources for YFCC15M-recap. The subset description is available at \url{https://github.com/openai/CLIP/blob/main/data/yfcc100m.md}. Images inherit the Creative Commons licenses of the original YFCC100M/Flickr media.

\item \textbf{DCI.} We use DCI for long-caption image-text retrieval evaluation. The dataset is available at \url{https://github.com/facebookresearch/DCI}
. DCI is released under the Creative Commons Attribution-NonCommercial 4.0 International License (CC-BY-NC 4.0), which permits reproduction, sharing, and adaptation for non-commercial purposes with attribution. We use DCI only for research purposes and follow its attribution and non-commercial use requirements. The dataset is provided ``AS IS'' without warranty. The associated images require acceptance of the SA-1B license agreement.

\item \textbf{DOCCI.} We use DOCCI for long-caption image-text retrieval evaluation. The dataset is available at \url{https://google.github.io/docci/}. The annotations and images are licensed by Google LLC under CC-BY-4.0, and we use the dataset following this license.

\item \textbf{ShareGPT4V.} We use ShareGPT4V-1k and ShareGPT4V-10k for long-caption image-text retrieval evaluation. The dataset is available at \url{https://sharegpt4v.github.io/}. The dataset is released under CC-BY-NC-4.0 and is intended for research use only. We use it only for research purposes and follow the associated license notices, including the restrictions related to CLIP, LLaMA, Vicuna, and GPT-4.

\item \textbf{Urban-1k.} We use Urban-1k for long-caption image-text retrieval evaluation. The dataset is available at \url{https://huggingface.co/datasets/BeichenZhang/Urban1k}. The Hugging Face dataset page specifies the license as CC-BY-NC-4.0, and we use it only for non-commercial research purposes.

\item \textbf{ImageInWords (IIW).} We use ImageInWords for long-caption image-text retrieval evaluation. The dataset is available at \url{https://google.github.io/imageinwords/}. The dataset is released under CC-BY-4.0, and we use it following the corresponding attribution requirements.

\item \textbf{RefCOCO, RefCOCO+, and GRef.} We use RefCOCO, RefCOCO+, and GRef for referring image segmentation evaluation. These datasets are accessed through the REFER API at \url{https://github.com/lichengunc/refer}. The REFER API is released under Apache-2.0. We cite the original RefCOCO, RefCOCO+, and GRef papers and follow the original dataset access terms. The associated images are from MS COCO and therefore follow the corresponding MS COCO/Flickr terms.

\item \textbf{MS COCO.} We use MS COCO as an image source for referring image segmentation evaluation and short-text retrieval evaluation. The dataset is available at \url{https://cocodataset.org/}. COCO annotations are released under CC-BY-4.0, while the images follow the licenses and terms of the original Flickr image owners. We use the dataset following the COCO annotation license and the applicable Flickr terms.

\item \textbf{Flickr.} Several datasets used in this work contain images originally hosted on Flickr. Flickr is available at \url{https://www.flickr.com/}. We follow the Flickr Terms of Use at \url{https://www.flickr.com/help/terms} and respect the image-level licenses specified by the original Flickr uploaders, including Creative Commons licenses where applicable.

\item \textbf{FLAIR.} Our implementation builds on the FLAIR codebase. The project repository is available at \url{https://github.com/ExplainableML/flair}
, and the released model is available at \url{https://huggingface.co/xiaorui638/flair}. The Hugging Face model card specifies the released model license as MIT. We cite the original FLAIR paper and acknowledge the FLAIR codebase as the basis for our implementation.

\end{itemize}

We cite the original papers or repositories for all existing assets and use them following their respective licenses and terms of use.

\section{The Use of Large Language Models (LLMs)}
\label{supp:llm}
Large Language Models (LLMs) were used solely as general-purpose editing tools to refine grammar and phrasing. They did not contribute to research ideation, analysis, or substantive writing.

\hfill



\newpage
\section*{NeurIPS Paper Checklist}

\begin{enumerate}

\item {\bf Claims}
    \item[] Question: Do the main claims made in the abstract and introduction accurately reflect the paper's contributions and scope?
    \item[] Answer: \answerYes{} 
    \item[] Justification: The claims made in the abstract and introduction are stated in Section \ref{sec:method} and Section \ref{sec:exp}.

\item {\bf Limitations}
    \item[] Question: Does the paper discuss the limitations of the work performed by the authors?
    \item[] Answer: \answerYes{} 
    \item[] Justification: We discuss the limitation of our work in \appx{supp:limitations}

\item {\bf Theory assumptions and proofs}
    \item[] Question: For each theoretical result, does the paper provide the full set of assumptions and a complete (and correct) proof?
    \item[] Answer: \answerNA{} 
    \item[] Justification: The paper does not include theoretical results or formal proofs.

    \item {\bf Experimental result reproducibility}
    \item[] Question: Does the paper fully disclose all the information needed to reproduce the main experimental results of the paper to the extent that it affects the main claims and/or conclusions of the paper (regardless of whether the code and data are provided or not)?
    \item[] Answer: \answerYes{} 
    \item[] Justification:  The paper provides sufficient details to reproduce the main experimental results, including model/training settings, and implementation details in \cref{sec:exp-setup} and \appx{supp:exp_setup}

\item {\bf Open access to data and code}
    \item[] Question: Does the paper provide open access to the data and code, with sufficient instructions to faithfully reproduce the main experimental results, as described in supplemental material?
    \item[] Answer: \answerYes{}  
    \item[] Justification: The code and pretrained models are publicly available on our GitHub and Hugging Face repositories (\appx{supp:release}).

\item {\bf Experimental setting/details}
    \item[] Question: Does the paper specify all the training and test details (e.g., data splits, hyperparameters, how they were chosen, type of optimizer) necessary to understand the results?
    \item[] Answer: \answerYes{} 
    \item[] Justification: The paper specify all training and test details in \appx{supp:exp_setup} and \cref{sec:exp}

\item {\bf Experiment statistical significance}
    \item[] Question: Does the paper report error bars suitably and correctly defined or other appropriate information about the statistical significance of the experiments?
    \item[] Answer: \answerNo{} 
    \item[] Justification: The paper evaluates the method on standard image-text retrieval benchmarks with fixed evaluation protocols and compares against prior work under the same settings. However, we do not provide error bars, confidence intervals, or statistical significance tests, so the statistical significance of the reported differences is not explicitly quantified.

\item {\bf Experiments compute resources}
    \item[] Question: For each experiment, does the paper provide sufficient information on the computer resources (type of compute workers, memory, time of execution) needed to reproduce the experiments?
    \item[] Answer:  \answerYes{} 
    \item[] Justification: The paper reports the compute resources and execution time required for the experiments in \appx{supp:exp_setup}.
    
\item {\bf Code of ethics}
    \item[] Question: Does the research conducted in the paper conform, in every respect, with the NeurIPS Code of Ethics \url{https://neurips.cc/public/EthicsGuidelines}?
    \item[] Answer: \answerYes{} 
    \item[] Justification: The research conducted in the paper conform with the NeurIPS Code of Ethics.

\item {\bf Broader impacts}
    \item[] Question: Does the paper discuss both potential positive societal impacts and negative societal impacts of the work performed?
    \item[] Answer: \answerYes{} 
    \item[] Justification: We discuss broader impacts of our research in \appx{supp:broader_impact}
    
\item {\bf Safeguards}
    \item[] Question: Does the paper describe safeguards that have been put in place for responsible release of data or models that have a high risk for misuse (e.g., pre-trained language models, image generators, or scraped datasets)?
    \item[] Answer: \answerYes{} 
    \item[] Justification: We discuss Safeguards of our model in \appx{supp:safe}

\item {\bf Licenses for existing assets}
    \item[] Question: Are the creators or original owners of assets (e.g., code, data, models), used in the paper, properly credited and are the license and terms of use explicitly mentioned and properly respected?
    \item[] Answer: \answerYes{} 
    \item[] Justification: We mention the licenses of the used assets in \appx{supp:licenses}.

\item {\bf New assets}
    \item[] Question: Are new assets introduced in the paper well documented and is the documentation provided alongside the assets?
    \item[] Answer: \answerYes{} 
    \item[] Justification:  The paper introduces new assets, including the trained model and code (\appx{supp:release}).

\item {\bf Crowdsourcing and research with human subjects}
    \item[] Question: For crowdsourcing experiments and research with human subjects, does the paper include the full text of instructions given to participants and screenshots, if applicable, as well as details about compensation (if any)? 
    \item[] Answer: \answerNA{} 
    \item[] Justification: The paper does not involve crowdsourcing or research with human subjects.

\item {\bf Institutional review board (IRB) approvals or equivalent for research with human subjects}
    \item[] Question: Does the paper describe potential risks incurred by study participants, whether such risks were disclosed to the subjects, and whether Institutional Review Board (IRB) approvals (or an equivalent approval/review based on the requirements of your country or institution) were obtained?
    \item[] Answer: \answerNA{} 
    \item[] Justification: The paper does not involve crowdsourcing nor research with human subjects.

\item {\bf Declaration of LLM usage}
    \item[] Question: Does the paper describe the usage of LLMs if it is an important, original, or non-standard component of the core methods in this research? Note that if the LLM is used only for writing, editing, or formatting purposes and does \emph{not} impact the core methodology, scientific rigor, or originality of the research, declaration is not required.
    \item[] Answer: \answerYes{} 
    \item[] Justification: Although LLMs were used only for grammar editing, or formatting purposes, we disclose their use in \appx{supp:llm}

\end{enumerate}

\end{document}